\documentclass[acmsmall]{acmart}

\acmJournal{TIST}

\usepackage{subfigure}
\usepackage{xspace}
\usepackage{multirow}
\usepackage{xcolor}
\usepackage{amsmath}
\usepackage{amsfonts}
\usepackage{bm}
\usepackage{soul}
\usepackage[ruled]{algorithm2e}
\usepackage{balance}
\usepackage{dsfont}
\usepackage[normalem]{ulem} % Part of the standard distribution
\newcommand{\rev}[1]{{\color{black}{#1}}}

\newcommand{\eat}[1]{}

\newcommand{\chengdu}{{Chengdu}\xspace}
\newcommand{\wuhan}{{Wuhan}\xspace}
\newcommand{\guangzhou}{{Guangzhou}\xspace}
\newcommand{\shaoxing}{{Shaoxing}\xspace}
\newcommand{\mianyang}{{Mianyang}\xspace}
\newcommand{\zhuhai}{{Zhuhai}\xspace}
\newcommand{\meta}{{MetaTransfer}\xspace}

\newcommand{\eg}{\emph{e.g.},\xspace}
\newcommand{\ie}{\emph{i.e.},\xspace}
\newcommand{\wrt}{\emph{w.r.t.}\xspace}

\newcommand\figref[1]{Figure~\ref{#1}}

\newcommand\tabref[1]{Table~\ref{#1}}
\newcommand\secref[1]{Section~\ref{#1}}

\newcommand\equref[1]{Eq.~(\ref{#1})}
  
\newcommand{\valstd}[2]{$#1 {\scriptstyle \,\pm\, #2}$}
\newcommand{\valstdb}[2]{$\textbf{#1} {\scriptstyle \,\pm\, #2}$}
\newcommand{\valstdu}[2]{$\underline{#1} {\scriptstyle \,\pm\, #2}$} 

\newtheorem{pro}{\textbf{Problem}}
\newtheorem{subpro}{\textbf{Sub-Problem}}
\newtheorem{defi}{\textbf{Definition}}

\begin{document}

\title{Meta-Transfer Learning Powered Temporal Graph Networks for Cross-City Real Estate Appraisal}

% \author{Weijia Zhang$^{1}$, Jindong Han$^{2}$, Hao Liu$^{1}$, Wei Fan$^{2}$, Hao~Wang$^{3}$, Hui Xiong$^{1}$}
% \affiliation{
% 	$^1$Hong Kong University of Science and Technology (Guangzhou),
% 	$^2$Hong Kong University of Science and Technology\\
% 	wzhang411@connect.hkust-gz.edu.cn,
% 	\{jhanao, wfanag\}@connect.ust.hk,
% 	\{liuh, xionghui\}@ust.hk
% }

\author{Weijia~Zhang}
\email{wzhang411@connect.hkust-gz.edu.cn}
\affiliation{%
  \institution{The Hong Kong University of Science and Technology~(Guangzhou)}
  \country{China}
}

\author{Jindong~Han}
% \email{jhanao@connect.ust.hk}
\email{jindong.han@sdu.edu.cn}
\affiliation{%
  \institution{Shandong University}
  \country{China}
}

\author{Hao~Liu}
\email{liuh@ust.hk}
\authornote{Corresponding author.}
\affiliation{%
\institution{The Hong Kong University of Science and Technology~(Guangzhou), The Hong Kong University of Science and Technology}
    \country{China}
}

\author{Wei~Fan}
\email{wfanag@connect.ust.hk}
\affiliation{%
\institution{The Hong Kong University of Science and Technology}
    \country{China}
}

\author{Hao~Wang}
\email{cashenry@126.com}
\affiliation{%
\institution{Computer Network Information Center, Chinese Academy of Sciences}
    \country{China}
}

\author{Hui Xiong}
\email{xionghui@ust.hk}
\authornotemark[1]
\affiliation{%
\institution{The Hong Kong University of Science and Technology~(Guangzhou), The Hong Kong University of Science and Technology}
    \country{China}
}

% \renewcommand{\shortauthors}{Weijia Zhang et al.}

\begin{abstract}
Real estate appraisal is important for a variety of endeavors such as real estate deals, investment analysis, and real property taxation. However, the market value of real estate can be simultaneously influenced by various complicated intrinsic and extrinsic factors, which impose great challenges on precise valuation.
Recently, deep learning has shown great promise for real estate appraisal by harnessing substantial online transaction data from web platforms. Nonetheless, deep learning is data-hungry, and thus it may not be trivially applicable to many small cities with limited data.
To this end, we propose Meta-Transfer Learning Powered Temporal Graph Networks~(\meta) to transfer valuable knowledge from multiple data-rich metropolises to the data-scarce city to improve valuation performance.
Specifically, by modeling the ever-growing real estate transactions with associated residential communities as a temporal event heterogeneous graph, we first design an Event-Triggered Temporal Graph Network to model the irregular spatiotemporal correlations among evolving real estate transactions.
Besides, we formulate the city-wide real estate appraisal as a multi-task dynamic graph link label prediction problem, where the valuation of each community in a city is regarded as an individual task.
A Hypernetwork-Based Multi-Task Learning module is proposed to simultaneously facilitate intra-city knowledge sharing between multiple communities and task-specific parameters generation to accommodate the community-wise real estate price distribution. 
Furthermore, we propose a Tri-Level Optimization Based Meta-Learning framework to adaptively re-weight training transaction instances from multiple source cities to mitigate negative transfer, and thus improve the cross-city knowledge transfer effectiveness.
\rev{Finally, extensive experiments based on six real-world datasets demonstrate the significant superiority of \meta compared with eleven baseline algorithms.}
\end{abstract}

\begin{CCSXML}
<ccs2012>
   <concept>
       <concept_id>10002951.10003227.10003236</concept_id>
       <concept_desc>Information systems~Spatial-temporal systems</concept_desc>
       <concept_significance>500</concept_significance>
       </concept>
 </ccs2012>
\end{CCSXML}

\ccsdesc[500]{Information systems~Spatial-temporal systems}

\keywords{Real estate appraisal; cross-city transfer learning; meta-learning; graph neural networks}

\maketitle

% \vspace{-3mm}
\section{Introduction}\label{sec:intro}

% Real estate appraisal aims to develop unbiased opinions of real property’s market value. It is a vital component of many real estate-related transactions, such as applying for a mortgage to purchase a property, refinancing the existing mortgage, or selling a property, especially when the buyer does not make an all-cash payment.
% In addition, real estate appraisal is also critical in extensive business and financial activities. The appraisal report provides a crucial tool for various stakeholders to make informed decisions about the property's worth and potential risks.
% For instance, it assists buyers or sellers in making decisions for property sales, directs investors in real estate investment, guides mortgage lenders to decide how much to lend, and helps the tax bureau to estimate property tax~\cite{zhao2019deep}. 
% Due to its significance and challenge, the real estate marketplace company even held over one million prize competition to improve its valuation capacities~\cite{sangani2017predicting}.
% Due to its significance and challenge, the real estate marketplace company even held over one million prize competition\footnotemark[1] to improve its valuation capacities~\cite{viana2021attention}.

% The significance and challenge of real estate appraisal have led to the efforts such as over one million prize competition\footnote[1]{https://www.kaggle.com/c/zillow-prize-1/} held by Zillow, a real estate marketplace company, to improve its valuation capabilities~\cite{viana2021attention}.

Real estate appraisal plays a crucial role in establishing fair and unbiased assessments of property market values, serving as a cornerstone in various real estate transactions~\cite{schram2006real,sayce2009real}. Whether applying for a mortgage, refinancing, or selling a property, accurate appraisals ensure equitable dealings and protect the interests of all parties involved. Beyond individual transactions, real estate appraisal serves the broader public good by underpinning key business and financial activities. Appraisal reports empower stakeholders, ranging from homebuyers and sellers to investors and lenders, to make well-informed decisions, manage risks, and ensure transparency in the market. Moreover, it assists tax authorities in accurately determining property taxes, thereby supporting the equitable distribution of public resources and contributing to the overall economic stability of society~\cite{pagourtzi2003real}.

\begin{figure}[tb]
  \centering
  % \hspace{-2mm}
%   \vspace{-1mm}
  \subfigure[{Distribution of real estate transactions}]{\label{fig:transvolume}
    \includegraphics[width=0.414\columnwidth]{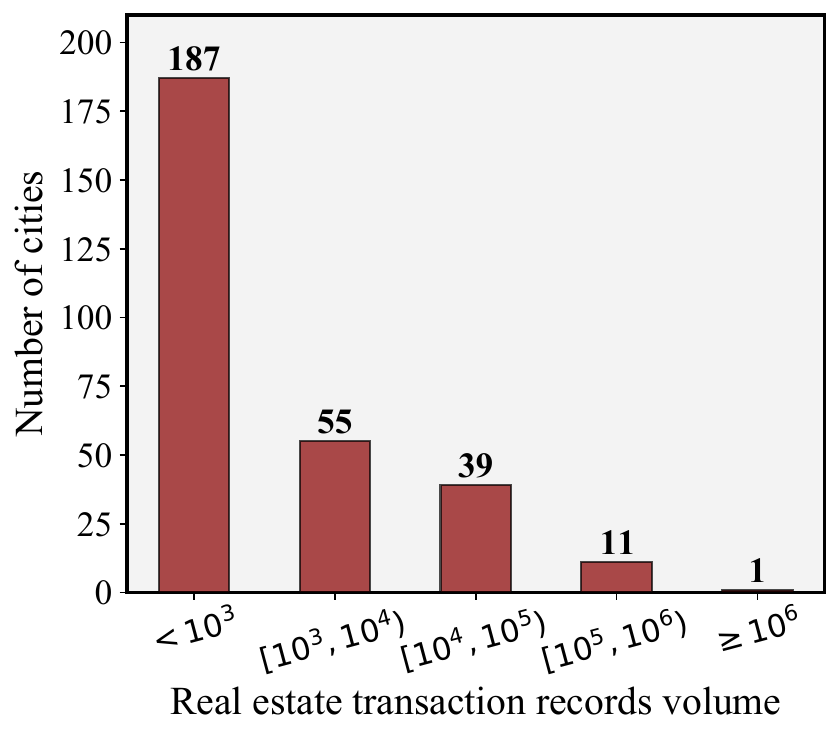}}
  \subfigure[{Cross-city real estate appraisal}]{\label{fig:transferproblem}
    \includegraphics[width=0.423\columnwidth]{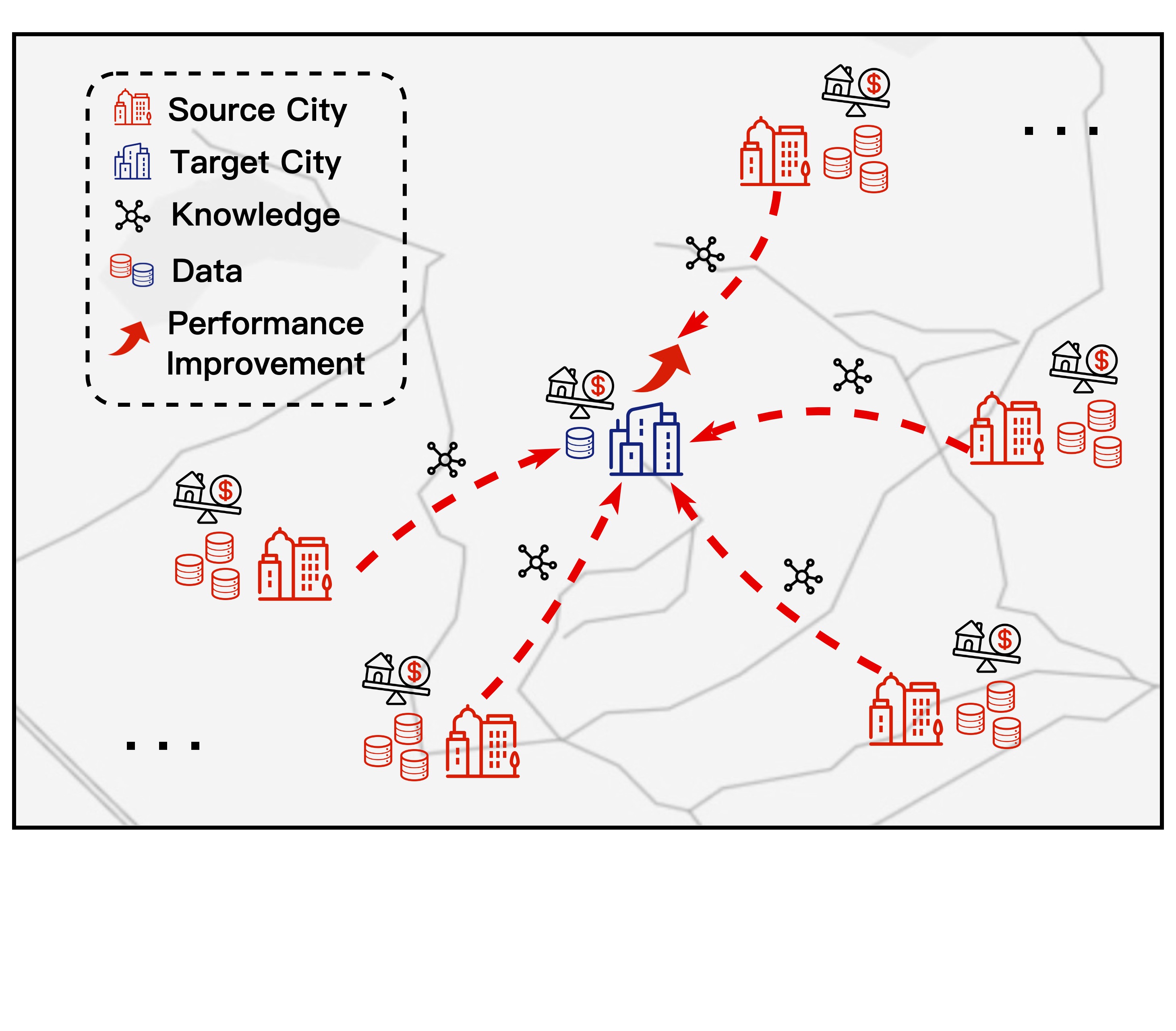}}
  \vspace{-3mm}
  \caption{(a)~Distribution of real estate transaction records volume of 293 Chinese cities, over 60\% of cities have fewer than 1,000 records. (b)~Illustration of the cross-city real estate appraisal task. The valuation performance in the data-scarce target city can be improved by transferring valuable knowledge from multiple data-rich source cities.} 
  \vspace{-3mm}
  \label{fig:problem}
\end{figure}

Existing real estate appraisal methods can be primarily classified as traditional valuation methods and automated valuation methods.
Traditional valuation methods, such as income approach, cost approach, sales comparison approach, and hedonic pricing approach, heavily rely on domain expertise, which hinders their prevalence among non-experts~\cite{pagourtzi2003real,zhang2021mugrep}. 
In contrast, automated valuation methods attempt to automatically generate valuations in a data-driven manner by leveraging machine learning algorithms, \eg~support vector regression~\cite{lin2011predicting} and boosted regression trees~\cite{park2015using}. 
Particularly, the emergence of web-based real estate platforms, such as \href{https://zillow.com}Zillow and \href{https://lianjia.com}Lianjia, has created an unparalleled opportunity to gather and utilize extensive online transaction data for automated property valuation.
This development has fueled the adoption of deep learning methods, which excel in leveraging these vast online data sources and capturing the complex, non-linear interactions among various real estate factors, leading to remarkable improvements in valuation accuracy~\cite{you2017image,law2019take,peter2020review, zhang2021mugrep,lee2023st}. 
These automated methods are easily accessible even to individuals outside the domain of expertise and thus exhibit remarkable practicability~\cite{niu2019intelligent}.

% \footnotetext[1]{https://www.kaggle.com/c/zillow-prize-1/}
% \footnotetext[1]{The statistical data is collected from one of the world’s largest commercial real estate agencies ``https://lianjia.com'' until September 2022.}
\footnotetext[1]{The statistical data is collected from one of the world’s largest commercial real estate web platforms ``https://lianjia.com'' until September 2022.}

Nevertheless, deep learning models are notorious for their data-hungry nature, in which large-scale training data is required to develop effective models~\cite{goodfellow2016deep,liu2021jizhi}.
\figref{fig:transvolume} presents statistics\footnotemark[1] of the web-sourced real estate transaction records volume across 293 Chinese prefecture-level cities.
As can be seen, over $60\%$ cities have less than 1,000 available real estate transaction records, which are far insufficient to train a deep learning model for satisfactory valuation performance.
Fortunately, abundant real estate transaction data has been collected from multiple metropolises, and extensive real estate market knowledge learned from these data-rich metropolises is also applicable to other data-scarce cities.
For example, no matter in which city, the real properties that have well-designed structures, fancy decorations, and are equipped with convenient living facilities around tend to have higher market values. 
It motivates us to investigate the cross-city real estate appraisal task as illustrated in \figref{fig:transferproblem}, which aims to improve the deep learning model's valuation performance in the data-scarce target city by transferring valuable knowledge from multiple data-rich source cities.

\rev{However, cross-city real estate appraisal is a non-trivial problem due to the following three major technical challenges: 
(1)~\emph{How to model the irregular spatiotemporal correlations among evolving real estate transactions?} 
The market value of real property is highly correlated to historical real estate transactions of geographical proximity.
However, these transactions are irregularly growing in spatial and temporal domains, which poses a challenge to model the irregular spatiotemporal correlations among evolving real estate  transactions and their price. 
% (2)~\emph{How to learn city-wise and community-wise knowledge simultaneously?}
(2)~\emph{How to collaboratively learn community-wise knowledge for each city?}
% As real estate market value is significantly influenced by its residential community~\cite{zhang2021mugrep}, it is useful to study real estate appraisal from the community perspective.
Different residential communities~\cite{lai2014outdoor} within a city can notably vary in terms of real estate prices.
Learning a completely shared valuation model for all communities fails to accommodate the community-wise real estate price distribution, whereas it is infeasible to learn an individually parameterized model for each community with sparse or even missing transaction data.
How to collaboratively learn the model to accommodate community-wise knowledge for each city is another challenge.
(3)~\emph{How to adaptively control the knowledge transfer from multiple source cities' transaction data to the target city?}
Each transaction instance from the source cities may exert a dynamically distinct effect on transferring knowledge to the target city in different training stages. For example, the knowledge distilled from some source cities' data instances may be helpful for the target city at the beginning of training, but the effect gradually deteriorates or even becomes negative as training proceeding.
How to adaptively adjust the effect of each source instance to mitigate negative knowledge transfer is the last challenge.
}

To tackle the above challenges, in this paper, we present Meta-Transfer Learning Powered Temporal Graph Networks~(\meta) for cross-city real estate appraisal.
Specifically, by modeling the ever-growing real estate transactions with the residential communities they occurred in as a temporal event heterogeneous graph and regarding real estate appraisal for each community of a city as an individual task, we first reformulate city-wide real estate appraisal as a multi-task dynamic graph link label prediction problem.
Then an Event-Triggered Temporal Graph Network is designed to model the irregular spatiotemporal correlations between the real estate price and evolving real estate transactions, where we sequentially incorporate the continuously growing transaction events of each community through a time-aware embedding evolution module and integrate the up-to-date real estate market knowledge of neighboring communities via a dimensional attentive graph convolution.
After that, we propose a Hypernetwork-Based Multi-Task Learning module to simultaneously facilitate intra-city universal knowledge sharing between multiple communities and task-specific parameters generation to accommodate the community-wise real estate price distribution.
Finally, by formulating each city as a task set, we propose a meta-learning method with instance re-weighting to adaptively transfer helpful knowledge from multiple source cities to the target city, and further propose a tri-level optimization framework to learn a weight-generating network for dynamic instance weights generation.

Our major contributions are summarized as follows:
(1)~We investigate a new cross-city real estate appraisal problem. To our knowledge, we are the first to explore how to transfer valuable knowledge from multiple data-rich source cities to the data-scarce target city to improve valuation performance. 
(2)~By reformulating city-wide real estate appraisal as a multi-task dynamic graph link label prediction problem, we design an Event-Triggered Temporal Graph Network to model the irregular spatiotemporal correlations of evolving real estate transactions and propose a Hypernetwork-Based Multi-Task Learning module to simultaneously facilitate intra-city knowledge sharing and community-wise knowledge learning.
(3)~We propose a Tri-Level Optimization Based Meta-Learning framework for effective cross-city knowledge transfer.
\rev{(4)~Extensive experiments on six real-world datasets demonstrate the effectiveness of \meta.}

% Our major contributions are summarized as follows:
% \begin{itemize}
% \item ~We investigate a new cross-city real estate appraisal problem. To our knowledge, we are the first to explore how to transfer valuable knowledge from multiple data-rich source cities to the data-scarce target city to improve valuation performance. 
% \item ~By reformulating city-wide real estate appraisal as a multi-task dynamic graph link label prediction problem, we design an Event-Triggered Temporal Graph Network to model the irregular spatiotemporal correlations of evolving real estate transactions and propose a Hypernetwork-Based Multi-Task Learning module to simultaneously facilitate intra-city knowledge sharing and community-wise knowledge learning.
% \item ~We propose a Tri-Level Optimization Based Meta-Learning framework for effective cross-city knowledge transfer.
% \item ~Extensive experiments based on six real-world datasets demonstrate the effectiveness of \meta.
% \end{itemize}

% The remainder of this paper is organized as follows: In \secref{sec:related}, we present the review of prior related works. In \secref{sec:preliminaries}, we describe the important definitions and formally formalize the cross-city real estate appraisal problem. \secref{sec:methodology} provides an overview of the proposed framework and then introduces its technical details. In \secref{sec:exp}, we report the experimental results based on six real-world datasets. Finally, we conclude this paper in \secref{sec:conclusion}.

% \vspace{-3mm}
% --PRELIMINARIES
\section{Preliminary}\label{sec:preliminaries}
\rev{In this section, we first introduce some important definitions of this work, then formally formulate the cross-city real estate appraisal problem.
% A city contains a set of residential communities $C=\{c_1,\cdots,c_{|C|}\}$, each of which is denoted as $c_i=\langle l_i, \mathbf{z}^t_i \rangle$, where $l_i$ is the location of $c_i$, and $\mathbf{z}^t_i$ is the attributes of $c_i$ at time $t$. 
% A residential community consists of a set of real estates, where each real estate $e$ has its own apartment attributes $\mathbf{x}$. 
% Please refer to \secref{sec:data_des} for the details of communities and apartment attributes.

\begin{defi}[\textbf{Real Estate}]
  Real estate refers to land together with any permanent structures or improvements attached to it, such as buildings or homes~\cite{jowsey2014real}. Each real estate $e$ has its own apartment attributes $\mathbf{x}$, such as house area, structure, decoration, and transaction ownership.
%   In general, real estate market value is influenced by both its apartment attributes $\mathbf{x}^k$ and attributes $\mathbf{z}_t_i$ of community the real estate located in.
\end{defi}

\begin{defi}[\textbf{Residential Community}]
  A residential community is an area or neighborhood primarily consisting of a group of housing real estates where people live~\cite{lai2014outdoor}. 
  A city comprises a set of residential communities $C=\{c_1,\cdots,c_{|C|}\}$, each of which is denoted as $c_i=\langle l_i, \mathbf{z}^t_i \rangle$, where $l_i$ is the location of $c_i$, and $\mathbf{z}^t_i$ is the attributes of $c_i$ at time $t$, such as its surrounding living facilities.
\end{defi}
}

\begin{defi}[\textbf{Transaction Event}]
% \noindent \emph{\textbf{Definition 1: Transaction Event.} 
% By regarding a real estate transaction as an event, 
We define the $n$-th chronological real estate transaction event in a city as $s_n =\langle e_n, c_i, t, y_n \rangle$ $\in S$, which indicates a real estate $e_n$ located in residential community $c_i$ is traded at a unit price of $y_n$ per square meter at time $t$, and $S$ denotes the set of transaction events in the city.
% }
\end{defi}

We further define a real estate $e_{\hat{n}}$ of community $c_i$ to be appraised its unit price $y_{\hat{n}}$ at time $\hat{t}$ as a potential transaction event $s_{\hat{n}}=\langle e_{\hat{n}}, c_i, \hat{t}, - \rangle $, where ``$-$'' denotes the unit price of transaction needs to be appraised. Then our cross-city real estate appraisal problem can be formally formulated as below:

\begin{pro}[\textbf{Cross-City Real Estate Appraisal}]
% \noindent \emph{\textbf{Problem 1: Cross-City Real Estate Appraisal.}
Given a set of source cities $U_{src} = \{u_1, ..., u_{|U_{src}|} \}$ with abundant historical real estate transaction data $\mathcal{D}_{src}$, a target city with scarce historical transaction data $\mathcal{D}_{tgt}$, our problem is to estimate real estate unit price $y_{\hat{n}}$ of each potential transaction event $s_{\hat{n}}=\langle e_{\hat{n}}, c_i, \hat{t}, - \rangle $ in the target city:
\begin{equation}
  f(s_{\hat{n}}; \mathcal{D}_{src}, \mathcal{D}_{tgt}) \rightarrow y_{\hat{n}},
\end{equation}
where $f(\cdot)$ is valuation model we aim to learn from $\mathcal{D}_{src}$ and $\mathcal{D}_{tgt}$.
% }
\end{pro}

% \vspace{-3mm}
% --Methodology
\section{Methodology}\label{sec:methodology}
% In this section, we first present an overview of the proposed framework, \meta. Then, we introduce the motivation and details to reformulate the city-wide real estate appraisal as a multi-task dynamic graph link label prediction problem. After that, we describe the technical details of using an Event-Triggered Temporal Graph Network to model the irregular spatiotemporal correlations between real estate transactions. Next, we detail the adoption of a Hypernetwork-Based Multi-Task Learning module to simultaneously facilitate intra-city knowledge sharing and community-wise knowledge learning. Furthermore, we elaborate on the proposed Tri-Level Optimization Based Meta-Learning framework for effective knowledge transfer from multiple source cities to a target city. 
% Finally, we analyze the complexity of the proposed model.

\begin{figure*}[tb]
  \centering
  % \hspace{-5mm}
  % \vspace{-2mm}
  \includegraphics[width=1\columnwidth]{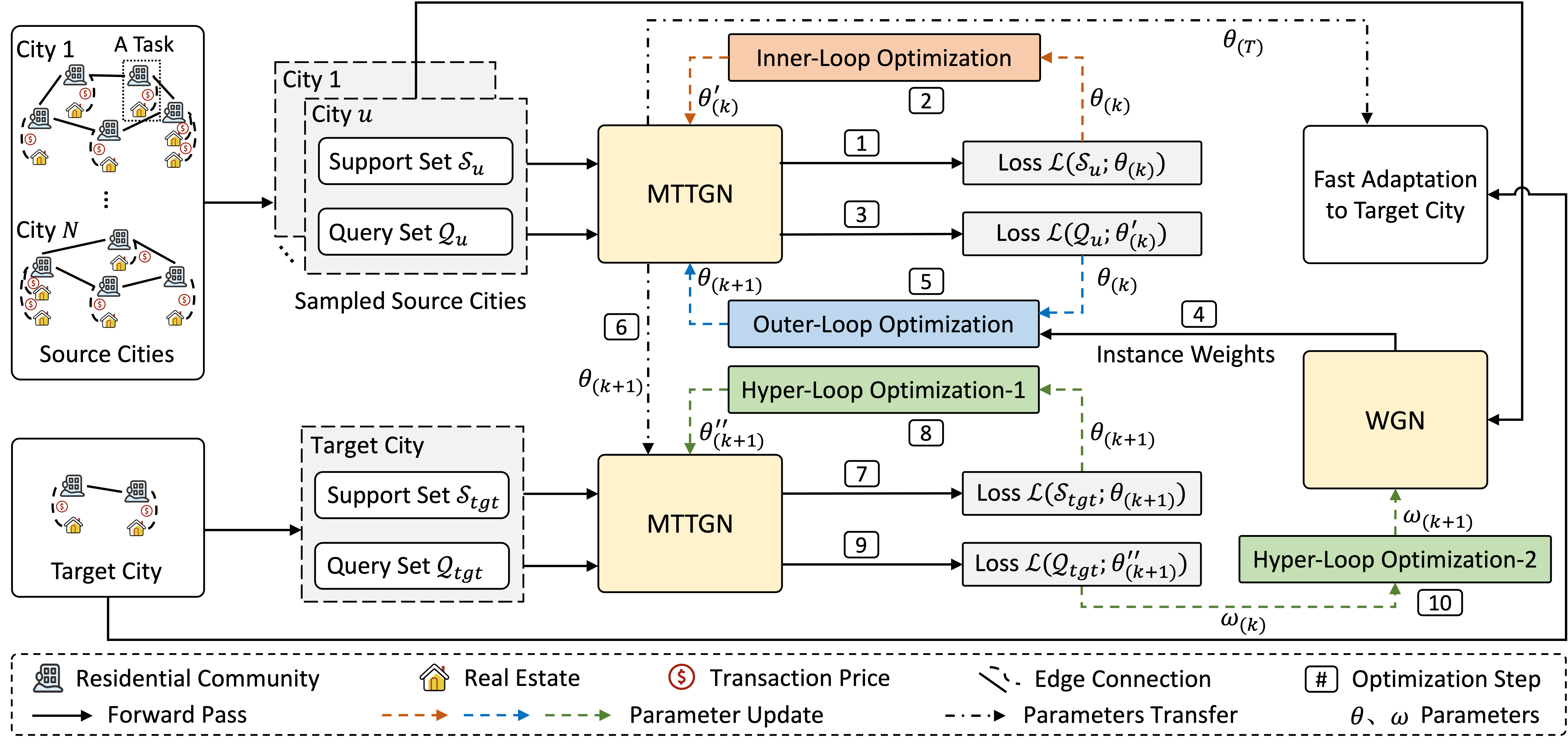}
  \vspace{-3mm}
  \caption{The framework overview of \meta. 
  % \TODO{refine this figure}
  % \TODO{This figure should be improved. First, what is the most important message you want to tell the audience? I think the loops and adaptions are more important than the loss and support query set. However, the current version lets the audience first see MTEGEN\&WGN(which is not explained in this figure, can you use some graphs and neural networks to make them more understandable?). As another example, use dollars rather than yuan, which may make the reviewers aware of where you are from.}
  } 
  % \vspace{-3mm}
  \label{fig:metatransfer}
\end{figure*}

% \begin{figure}[tb]
%   \centering
%   % \hspace{-5mm}
%   % \vspace{-2mm}
%   \includegraphics[width=1.05\columnwidth]{figs/metatransfer.png}
%   \vspace{-4mm}
%   \caption{The framework overview of \meta.} 
%   \vspace{-4mm}
%   \label{fig:metatransfer}
% \end{figure}

% \vspace{-1mm}
\subsection{Framework Overview}
\figref{fig:metatransfer} shows the framework overview of \meta. By modeling each city's residential communities with the real estate transactions as a temporal event heterogeneous graph and regarding the real estate appraisal of each community as an individual task, the city-wide real estate appraisal is reformulated as a multi-task dynamic graph link label prediction problem.
Then, we employ a Multi-Task Temporal Graph Network~(MTTGN), which consists of an Event-Triggered Temporal Graph Network and a Hypernetwork-Based Multi-Task Learning module, as the base model for real estate appraisal. 
% MTTGN consists of a Temporal Graph Network to model the irregular spatiotemporal correlations among evolving real estate transactions, and a Hypernetwork-Based Multi-Task Learning module to simultaneously facilitate intra-city universal knowledge sharing between communities and task-specific parameters generation to accommodate the community-wise real estate price distribution. 
In each meta-training iteration, a set of source cities are first sampled. For each sampled city, the data is split as a support set and a query set for meta-transfer learning. Then, the inner-loop and outer-loop optimizations are sequentially performed to distill source cities' knowledge into MTTGN's parameters, where a set of instance weights are generated by a Weight-Generating Network~(WGN) to re-weight each source instance's gradient during outer-loop optimization for mitigating negative knowledge distillation.
Next, we introduce an extra hyper-loop optimization to update WGN's parameters by evaluating MTTGN on target city's available data.
After multiple meta-training iterations are completed, MTTGN's parameters that contain extensive real estate market knowledge will be used to initialize the model for fast adaptation to target city.

% We first model each city as a temporal event heterogeneous graph, where nodes indicate real estates and residential communities, edges represent transaction events and communities' spatial proximity. By regarding the real estate appraisal of each community in a city as individual task, we formulate city-wide real estate appraisal as a multi-task dynamic graph link label prediction problem. Then we devise a \emph{Temporal Graph Network} to jointly incorporate the spatial correlations between communities and past-current temporal correlations of real estate transactions. After that, we introduce a \emph{Hypernetworks-Based Multi-Task Learning} framework to facilitate intra-city knowledge sharing and model the unique real estate price distribution of each community. Finally, we develop a meta-learning module that eliminates negative source knowledge and transfers helpful knowledge from multiple source cities to the target city. Specifically, the meta-learning module consists of a \emph{weight-generating network} to adaptively re-weight transaction instances from source cities and a \emph{Tri-Level Optimization} strategy to perform joint meta-training between source cities and target city.

% \vspace{-3mm}
\subsection{Multi-Task Dynamic Graph Link Label Prediction Formulation}
% As motivated, the prices of real estate transactions are highly correlated in both spatial and temporal domain, \ie transactions that are spatially and temporally adjacent tend to have a similar price. Moreover, the transaction price also exhibits strong dependencies with its associated residential community~\cite{fu2015real}. To characterize the complicated spatiotemporal relationships, we first construct a temporal event heterogeneous graph, defined as
Real estate market value is decided not only by its apartment attributes but also by the attributes of the community it belongs to~\cite{fu2015real}. For example, real estate located in a community with complete facilities around, \eg malls, subway stations, and schools, tend to have a higher price than belonging to a community in a desolate place. Furthermore, the real estate prices are highly correlated with historical real estate transactions that occurred in the same community and neighboring communities~\cite{zhang2021mugrep}. 
However, as transaction events can continually occur at any location~(community) and time, they are exhibiting irregular growth patterns in both spatial and temporal domains. This poses a significant challenge to model the irregular spatiotemporal correlations between real estate prices and evolving transaction events.
Therefore, we construct a temporal event heterogeneous graph to effectively characterize the above irregular spatiotemporal correlations, defined below:

\begin{defi}[\textbf{Temporal Event Heterogeneous Graph}]
% \noindent \emph{\textbf{Definition 2: Temporal Event Heterogeneous Graph.}
A temporal event heterogeneous graph is defined as $\mathcal{G}_t=(\mathcal{V}_t,\mathcal{E}_t)$, where $t$ denotes the time when a transaction event occurs, $\mathcal{V}_t$ is a set of heterogeneous nodes including real estates and residential communities at $t$, and $\mathcal{E}_t$ is a set of undirected edges indicating connectivity between nodes at $t$.
% }
\end{defi}

Due to the node heterogeneity in the graph, we define two types of edges, $e$-$c$ and $c$-$c$. Specifically, $e$-$c$ edges can be represented as $\langle e,c,t,y \rangle$ describing a transaction event that occurred to real estate $e$ located in the community $c$ at time $t$, whereas $c$-$c$ edges reflect the spatial proximity between two communities. 
% Thus, the graph is continually updated via a sequence of time-ordered transaction events. 
Thus, the graph is continually evolving with new transaction events occurring, or new communities joining.
We first define the edge connection between a real estate $e_n$ and a community $c_i$ when the transaction event $s_n =\langle e_n, c_i, t, y_n \rangle$ occurs:
% \vspace{-1mm}
\begin{equation}
e_n\text{-}c_i=\left\{
\begin{array}{lr}
\text{True},\quad\;\text{if}~s_n~occurred\\
\text{False},\quad \text{if}~otherwise
\end{array}
\right.,
% \vspace{-1mm}
\end{equation}
The edge connection between community $c_i$ and community $c_j$ is defined as:
% \vspace{-1mm}
\begin{equation}
c_i\text{-}c_j=\left\{
\begin{array}{lr}
\text{True},\quad\;\text{if}~p_{ij} < \epsilon\\
\text{False},\quad \text{if}~otherwise
\end{array}
\right.,
% \vspace{-1mm}
\end{equation}
where $p_{ij}$ is the spherical distance between $c_i$ and $c_j$, $\epsilon$ is a distance threshold, and each community has a self-loop to connect with itself.
In particular, each $e$-$c$ edge is associated with a label $y$, indicating the price of the corresponding transaction. Hence, given the real estate node, community node, and transaction time of their edge, real estate appraisal can be reformulated as a link label prediction problem on the temporal event heterogeneous graph.

% \TODO{Aware widow word!}
% Specifically, the graph $\mathcal{G}^t$ can be formed via sequentially updating an initial graph $\mathcal{G}^{t_0}$ according to the a sequence of time-ordered transaction events $S=\{s_{t_1},s_{t_2},\dots,s_{t_T},\dots\}$ that occurred before $t$.

\begin{figure}[tb]
  % \vspace{-4.5mm}
  \centering
  % \vspace{-2.5mm}
  % \hspace{-2mm}
  \subfigure[{Communities in Guangzhou}]{
    \includegraphics[width=0.44\columnwidth]{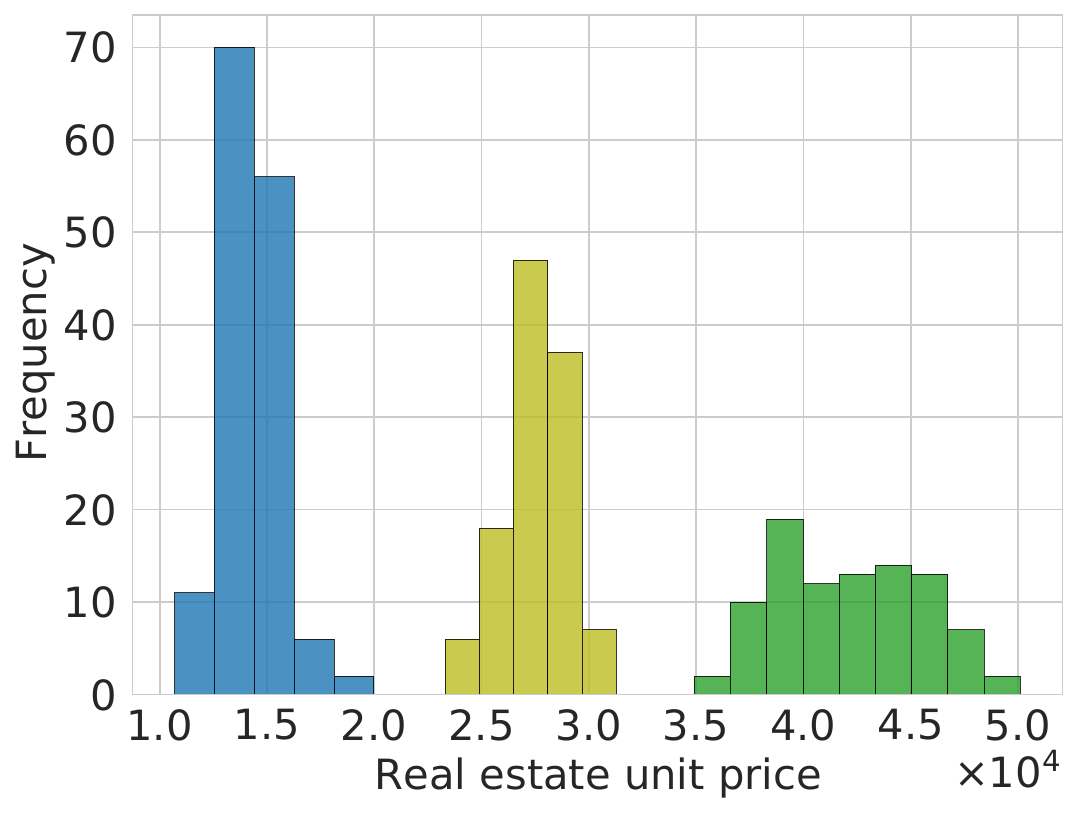}}
  \subfigure[{Communities in Mianyang}]{
    \includegraphics[width=0.43\columnwidth]{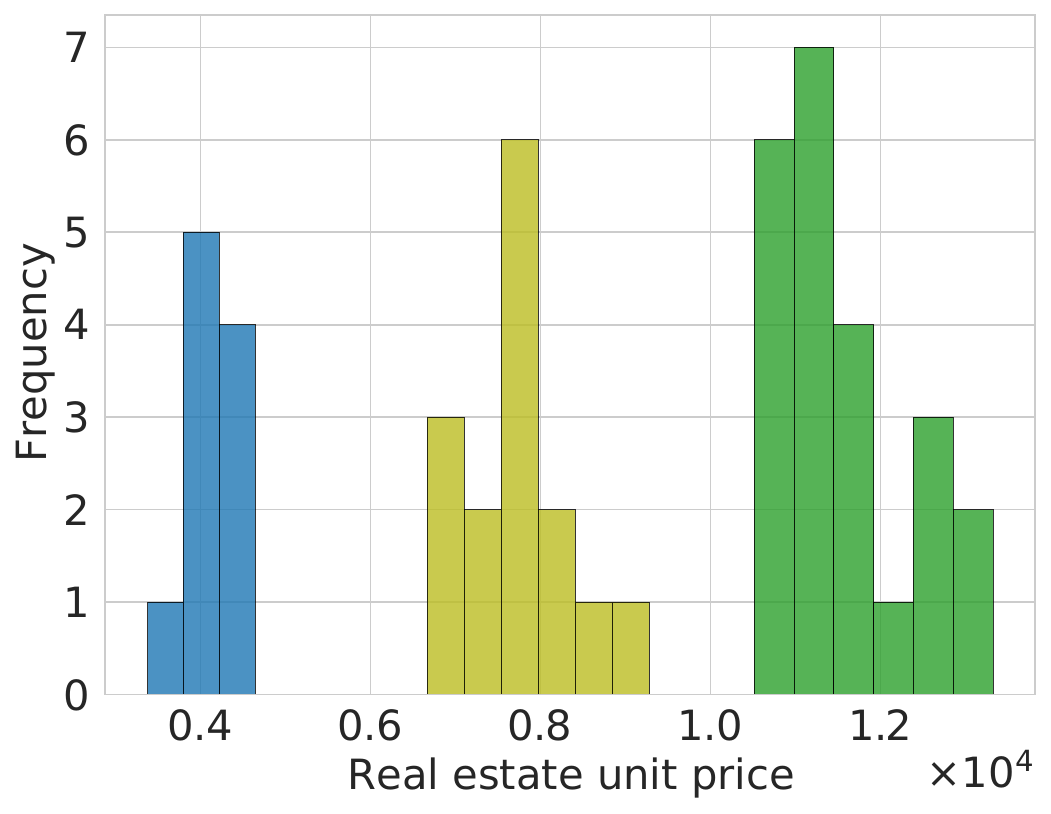}}
  \vspace{-3mm}
  \caption{Illustrations of the diversity between real estate price distributions of communities from cities Guangzhou and Mianyang, where different communities in a city are denoted by different colors.} 
  % \vspace{-5mm}
  \label{fig:community_price}
\end{figure}

\begin{figure}[tb]
  \centering
  \includegraphics[width=0.8\columnwidth]{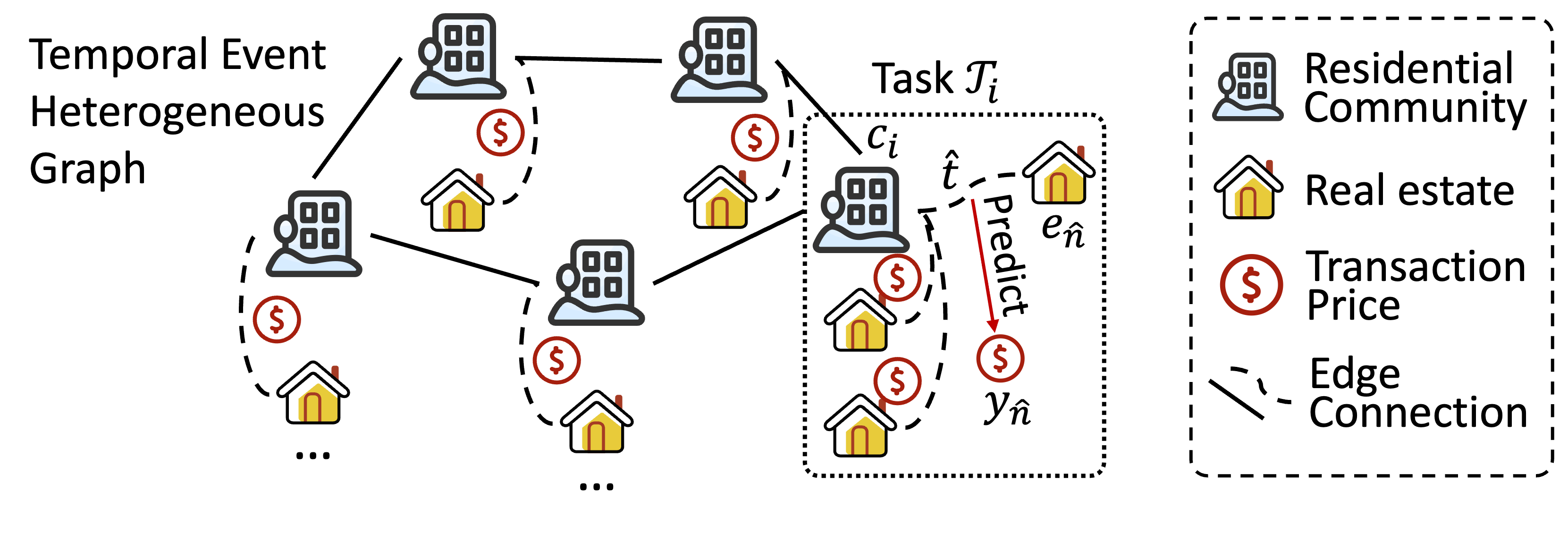}
  \caption{City-Wide Multi-Task Dynamic Graph Link Label Prediction Problem.} 
  \label{fig:subproblem}
\end{figure}

Besides the spatiotemporal correlations, as shown in \figref{fig:community_price}, we observe the apparently diverse distributions of real estate prices in different communities.
% depending on their own characteristics and surrounding environments. 
To handle such diversity, we regard the real estate appraisal for each community as an individual task $\mathcal{T}_i$ and real estate transactions that occurred in a community as the instances of a task. Then, as illustrated in \figref{fig:subproblem}, we further reformulate the real estate appraisal for each city as a city-wide multi-task dynamic graph link label prediction sub-problem:
% Formally, consider a set of tasks $\{\mathcal{T}_i\}_{i=1}^{|C|}$ at a city $u$, where $\mathcal{T}_i$ corresponds to the real estate appraisal task of a community $c_i$, the sub-problem of each city is defined as below.

\begin{subpro}[\textbf{City-Wide Multi-Task Dynamic Graph Link Label Prediction}]
% \noindent \emph{\textbf{Sub-Problem 1: City-Wide Multi-Task Dynamic Graph Link Label Prediction.}
Given a temporal event heterogeneous graph $\mathcal{G}_t$, a set of tasks ${\{\mathcal{T}_i\}}_{i=1}^{|C|}$, a real estate $e_{\hat{n}}$ to be appraised and the residential community $c_i$ it belongs to, the sub-problem for each city is to predict the link label, \ie transaction unit price, $y_{\hat{n}}$ between $e_{\hat{n}}$ and $c_i$ at time $\hat{t}$ under the task $\mathcal{T}_{i}$.
% }
\end{subpro}

% \vspace{-3mm}
\subsection{Irregular Spatiotemporal Correlations Modeling Between Transactions}
\begin{figure}[tb]
  \centering
  \includegraphics[width=0.98\columnwidth]{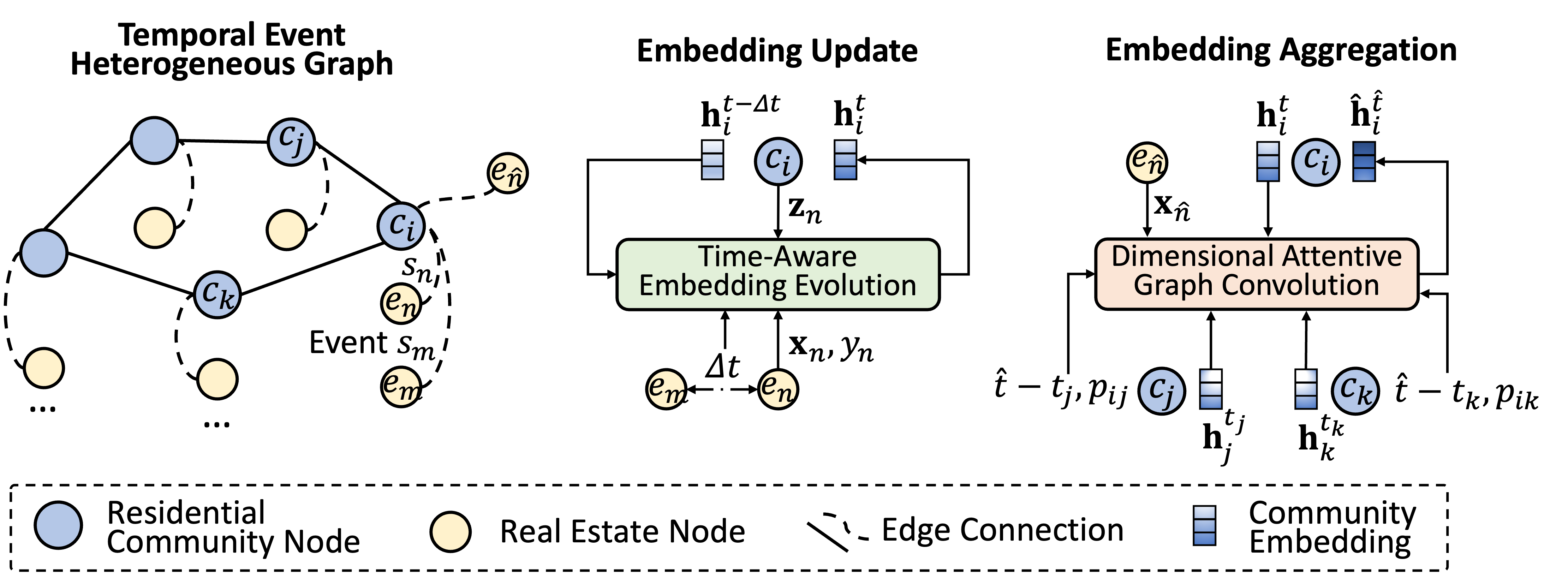}
  \caption{Event-Triggered Temporal Graph Network. It encompasses a time-aware embedding evolution module to incorporate evolving transaction events for the embedding update and a dimensional attentive graph convolution to integrate neighboring nodes for the embedding aggregation.}
  \label{fig:tgn}
\end{figure}

Since transaction events may occur at any time in a community and the spatial connections between communities are also irregular, which induces both temporal and spatial irregularity in the temporal event heterogeneous graph.
To this end, we design an Event-Triggered Temporal Graph Network~(TGN) to capture irregular spatiotemporal correlations based on the temporal event heterogeneous graph. 
As illustrated in \figref{fig:tgn}, TGN enables to incorporate the evolving transaction events to sequentially update the embedding of communities involved in the events through a time-aware embedding evolution module and also integrate the up-to-date embedding of neighboring communities via a dimensional attentive graph convolution.
% We describe the details below.

% We first initialize the state of each community and estate as its original features.
% After that, 

\subsubsection{\textbf{Time-Aware Embedding Evolution}}
% \noindent \textbf{Embedding Update.}
To incorporate the time-varying real estate market knowledge from each community, we maintain a temporal state embedding $\mathbf{h}_{i}^{t}$ for each community $c_i$ by dynamically updating it along with the occurrence of transaction event $s_n =\langle e_n, c_i, t, y_n \rangle$. 
% For simplicity, we will drop the superscript $i$ and $j$ in the following section.
While the time elapsing between two successive transaction events in a community usually varies, these time intervals hold valuable insights into the volatility and cyclicality of the real estate market~\cite{born1994real}.
% To effectively model the temporally irregular transaction events, we further extend Gated Recurrent Unit (GRU)~\cite{chung2014empirical}, a simple yet effective variant of recurrent neural networks~\cite{liu2020multi}, to a Time-Aware GRU.
Therefore, we develop a time-aware embedding evolution module to update the community's embedding by effectively capturing the knowledge from temporally irregular transaction events and carefully considering the time interval between events.
Formally, given a transaction event $s_{n} =\langle e_n, c_i, t, y_n \rangle$ and the last state embedding $\mathbf{h}_i^{t-\Delta t}$ of $c_i$, where $t-\Delta t$ represents the time of last transaction involving $c_i$ before $s_n$ and $\Delta t$ is the time interval, we adopt a time gating mechanism to control the influence of a community's past knowledge:
% \vspace{-1mm}
\begin{equation}
\mathbf{g}^{t-\Delta t}_i=\exp \left(-\operatorname{ReLU}\left(\mathbf{W}_g\ [\phi(\Delta t) \mathbin\| \mathbf{x}_n \mathbin\| y_n \mathbin\| \mathbf{h}^{t-\Delta t}_i ]\right)\right),
% \vspace{-3mm}
\end{equation}
\begin{equation}
\widetilde{\mathbf{h}}_{i}^{t-\Delta t}=\mathbf{g}_i^{t-\Delta t} \odot \mathbf{h}_i^{t-\Delta t},
% \vspace{-1mm}
\end{equation}
where $\mathbin\|$ indicates the concatenation operation,
% $\mathbf{x}^k$ and $y_ij$ denotes the features of real estate $e^k$ in transaction event $s_t$,
$\mathbf{W}_g$ and $\operatorname{ReLU}(\cdot)$ are learnable parameters and activation function, respectively, and $\odot$ denotes element-wise product.
To learn to characterize real estate market's cyclicality within the time interval, we adopt the original time interval with learnable Fourier features~\cite{xu2020inductive} to derive the time embedding:
% \begin{equation}
% \begin{aligned} 
% \phi(\Delta t) = & \left[\Delta t,\cos (w_{1} \Delta t), \sin (w_{1} \Delta t), \ldots, \cos (w_{d_{\phi}} \Delta t),\right. \\
% & \left. \sin (w_{d_{\phi}} \Delta t)\right],
% \end{aligned}
% \end{equation}
\begin{equation}
\phi(\Delta t) = \left[\Delta t,\cos (w_{1} \Delta t), \sin (w_{1} \Delta t), \ldots, \cos (w_{d_{\phi}} \Delta t), \sin (w_{d_{\phi}} \Delta t)\right],
\end{equation}
where $w_{1}, \ldots, w_{d_{\phi}}$ are learnable parameters. Then we update the temporal state embedding to $\mathbf{h}^{t}_i$ via a Gated Recurrent Unit~(GRU) operation~\cite{chung2014empirical}:
\begin{equation} 
\mathbf{h}_i^{t}=\operatorname{GRU}\left([\mathbf{x}_n \mathbin\| y_n \mathbin\| \mathbf{z}^t_i],\ \widetilde{\mathbf{h}}_{i}^{t-\Delta t}\right).
\end{equation}

\subsubsection{\textbf{Dimensional Attentive Graph Convolution}}
% \noindent {\textbf{Embedding Aggregation.}}
% After obtaining the dynamic state of $\mathbf{h}_i^{t}$ for community $c_i$, we can directly combine $\mathbf{h}_i^{t}$ with real estate features $\mathbf{x}^k$ of a target estate located in $c$ for subsequent link label prediction at a particular future time $t+\Delta t$.
Given a real estate $e_{\hat{n}}$ to be appraised and the latest state embedding $\mathbf{h}_i^{t}$ of community $c_i$ it belongs to, we can directly combine the features $\mathbf{x}_{\hat{n}}$ of real estate node and embedding $\mathbf{h}_i^{t}$ of community node with the time $\hat{t}$ of a potential transaction to predict the link label, namely the real estate unit price $y_{\hat{n}}$.
However, the embedding of a community is updated only when a transaction event occurs there. If a community does not have transaction events for a long time, the community embedding $\mathbf{h}_i^{t}$ representing past real estate market knowledge becomes stale, which may induce a negative effect on real estate appraisal in the current market environment. In addition, it fails to model the spatial correlation of real estate transactions.

% However, since the embedding of a community is updated only when there is a transaction event occurred, the same results will be made regardless of which future time point we are going to predict. As an illustration, consider that a community does not have any transaction activities for a long time. Now we need to estimate the price of real estate located in this community, the real estate price may have already changed a lot. Thus the community embedding will become stale.

To tackle the above problems, inspired by the recent success of Graph Neural Networks~(GNNs) on graph modeling~\cite{velivckovic2017graph,hamilton2017inductive,jiang2022graph}, we devise a dimensional attentive graph convolution to refresh community embedding by adaptively integrating the up-to-date knowledge from neighboring communities. 
Specifically, as knowledge from different dimensional features of a neighboring community may have distinct effects, we first introduce a feature-wise attention operation to derive dimensional correlation weights $\bm{w}_{ij} \in \mathbb{R}^{d}$:
% time-varying embedding of a community $c_i$ by aggregating embedding of its neighboring communities
\begin{equation}
\begin{gathered}
\bm{w}_{ij}=\frac{\exp(\bm{a}_{ij})}{\sum_{c_k\in\mathcal{N}_i} \exp(\bm{a}_{ik})}, \\
\bm{a}_{ij}=\mathbf{W}_2\ \sigma \left(\mathbf{W}_1\ [\mathbf{x}_{\hat{n}} \mathbin\| \mathbf{h}_i^t \mathbin\| \mathbf{h}_j^{t_j} \mathbin\| \phi(\hat{t}-t_j) \mathbin\| p_{ij}]\right),
% \bm{\beta}_{ij}=\mathbf{W}_{\beta}\tanh \left( \bm{\mathcal{H}}_{ij} \right),\\
% \bm{\mathcal{H}}_{ij} = \mathbf{W}_1[\mathbf{x}_{\hat{n}}  \mathbin\| \mathbf{h}_i^t \mathbin\| z_i^{\hat{t}}] + \mathbf{W}_2[\mathbf{h}_j^{t_j} \mathbin\| z_j^{\hat{t}}] + \mathbf{W}_3[\phi(\hat{t}-t_j) \mathbin\| p_{ij}],
\end{gathered}
% \vspace{-1mm}
\label{equ:weight}
\end{equation}
where $\mathcal{N}_i$ represents the neighboring communities of $c_i$ (including $c_i$ itself), $t_j$ denotes the last update time of $\mathbf{h}_j^{t_j}$ when recent transaction event occurred in $c_j$, $\sigma$ represents an activation function, $\mathbf{W}_1\in \mathbb{R}^{d_{in} \times d}$,  $\mathbf{W}_2 \in \mathbb{R}^{d \times d}$ are learnable parameters, $d_{in}$ and $d$ are feature dimensions, and $\bm{a}_{ij}\in \mathbb{R}^{d}$ is the attention score between $c_i$ and $c_j$.
Note we also consider edges features influence, \ie time interval $\phi(\hat{t}-t_j)$ and distance $p_{ij}$, for correlation weights computation. With $\bm{w}_{ij}$ obtained, the embedding of community $c_i$ is refreshed by aggregating up-to-date embeddings of its neighbors:
% \vspace{-1mm}
\begin{equation} 
{\widehat{\mathbf{h}}_i^{\hat{t}}}= \sigma\left(\mathbf{W}_{3} \sum_{c_j \in \mathcal{N}_i} \bm{w}_{ij} \odot \mathbf{h}_j^{t_j}\right),
% \vspace{-1mm}
\label{equ:aggregate}
\end{equation}
where $\mathbf{W}_3$ are learnable parameters.
% Through the above operation, even if a community has been inactive for a long period of time, we can still derive up-to-date embedding by aggregating its active neighbors.

% \vspace{-3mm}
\subsection{Intra-City Knowledge Sharing}
% Previously, we have formulated real estate appraisal as a multi-task learning problem by regarding real estate appraisal of each community as a distinct task.
As aforementioned, each community in a city is formulated as an individual task to meet unique real estate price distribution. 
Directly learning individual parameters for each community is infeasible because of limited or even missing transaction data of communities. 
Indeed, extensive universal knowledge hidden in real estate transactions can be shared by all tasks of a city.
To enforce intra-city knowledge sharing and simultaneously capture task-specific knowledge, a straightforward method is to jointly learn a shared feature extractor for all tasks in a city, while separately learning the task-specific output layer.
Nonetheless, the above approach is inappropriate for our problem setting due to the following three reasons. 
% First, the number of training instances in different tasks are highly unbalanced, 
First, many tasks only have extremely sparse transaction data or even no data available, which cannot support task-specific parameter learning.
% These tasks cannot benefit from such strategy due to data scarcity. 
Second, there are thousands of residential communities in a city, and training task-specific output layers for such massive tasks may suffer from low learning efficiency and poor scalability.
Third, it is difficult to generalize task-specific parameters from a task set~(\ie~a set of tasks) in the source city to another task set in the target city, which degrades model's capability in cross-city knowledge transfer.

To overcome the above limitations, we propose a Hypernetwork-Based Multi-Task Learning~(HMTL) module. The key idea is to generate task-specific output layer's parameters based on a learnable hypernetwork~\cite{ha2016hypernetworks} shared across all tasks of a city. 
Concretely, the hypernetwork takes community attributes as input to capture each task's unique characteristic, then generates the personalized output layer's parameters for each task via the following operations:
% \vspace{-3mm}
\begin{equation}
\begin{gathered}
\mathbf{W}_i^{\Phi}=\Phi_W(\mathbf{z}^{\hat{t}}_i),\\
\mathbf{b}_i^{\Phi}=\Phi_b(\mathbf{z}^{\hat{t}}_i),
\end{gathered}
% \vspace{-1mm}
\end{equation}
where $\Phi_W(\cdot)$ and $\Phi_b(\cdot)$ are the hypernetworks, which can be instantiated by learnable neural networks~(\eg~Multi-Layer Perceptron~(MLP)).
Once each task's generated parameters are obtained, the real estate $e_{\hat{n}}$ located in community $c_i$ can be appraised based on feature embeddings of nodes and time associated with the link:
\begin{equation}
\widehat{y}_{\hat{n}}=\mathbf{W}_i^{\Phi} \cdot \mathcal{F}\left([\mathbf{x}_{\hat{n}} \mathbin\| \mathbf{z}^{\hat{t}}_i \mathbin\| \widehat{\mathbf{h}}_i^{\hat{t}} \mathbin\| \hat{t}] \right)+\mathbf{b}_i^{\Phi},
\label{equ:appraisal}
\end{equation}
where $\mathcal{F}(\cdot)$ denotes a learnable feature integration function instantiated by a MLP.
In this way, tasks with similar community attributes will derive similar generated parameters, thus universal knowledge can be further shared across these tasks, which effectively alleviates data scarcity issue for tasks with sparse or even no instances. 
In addition, since the hypernetwork is shared by all tasks, the learnable parameters are independent of the number of tasks, which largely improves the model's learning efficiency and scalability on massive tasks, and the learned hypernetwork can be easily generalized to a new task set in the target city.

% Overall, given a task set in a city, our model aims to jointly minimize the Mean Absolute Error~(MAE) loss between the estimated unit price and ground truth unit price of real estate transactions for all tasks:
% \begin{equation}
% \mathcal{L}(S; \theta)=\frac{1}{|S|} \sum_{\langle e_{\hat{n}}, c_i, \hat{t}, y_{\hat{n}} \rangle \in S} \left|\ \widehat{y}_{\hat{n}}-y_{\hat{n}}\ \right|,
% \label{method:loss}
% \end{equation}
% where $\theta$ denotes all learnable parameters in TGN and HMTL.

% \vspace{-3mm}
\subsection{Cross-City Knowledge Transfer}
% The proposed Temporal Graph Network and hypernetwork are supervised models for real estate appraisal, which requires abundant transaction data to achieve satisfactory performance.
We elaborate the proposed Tri-Level Optimization Based Meta-Learning framework for effective knowledge transfer from multiple source cities to a target city.
The framework is comprised of two components: meta-learning with instance re-weighting and tri-level optimization.

\subsubsection{\textbf{Meta-Learning with Instance Re-Weighting}}
% \noindent{\textbf{Meta-Learning with instance re-weighting.}}
Transfer learning has been introduced as a powerful technique for cross-city knowledge transfer~\cite{wei2016transfer, wang2019cross, ijcai2022p282}. However, the above studies using typical transfer learning methods are limited to transferring knowledge from only a single source city, which causes a risk of unstable and useless transfer~\cite{yao2019learning}.
Inspired by the recent success of meta-learning in handling data-scarce traffic prediction problems by transferring knowledge from multiple source cities~\cite{yao2019learning,luCrossCityTransfer22}, we recast the cross-city knowledge transfer for real estate appraisal as a meta-learning problem.
As aforementioned, we formulate each city as a task set for city-wide multi-task learning.
\rev{Therefore, different from prior studies~\cite{yao2019learning,luCrossCityTransfer22} that introduce meta-learning to transfer knowledge from multiple tasks extracted from source cities to a new task of target city, we aim to develop meta-learning to extract transferable knowledge from multiple task sets~(multiple source cities) with abundant transaction data to rapidly adapt such knowledge to a new task set~(a target city) with insufficient data.}

Specifically, given a task set in a city, the base model MTTGN aims to jointly minimize the Mean Absolute Error~(MAE) loss between the estimated unit price and ground truth unit price of real estate transactions for all tasks:
\begin{equation}
\mathcal{L}(S; \theta)=\frac{1}{|S|} \sum_{s_n \in S} \left|\ \widehat{y}_{{n}}-y_{{n}}\ \right|,
\label{method:loss}
\end{equation}
where $\theta$ denotes all learnable parameters in MTTGN, and are exactly the knowledge to be generalized to the target city.

Analogous to the bi-level optimization in MAML~\cite{finn2017model}, we first sample a batch of source cities from $U_{src}$. For each sampled city $u$, we split the transaction data into a support set $\mathcal{S}_{u}$ and a query set $\mathcal{Q}_{u}$. 
We denote $\theta_{(k)}$ as model parameters after $k$-th iterations.
In the inner-loop optimization, we adapt the model parameters $\theta_{(k)}$ on support set:
\begin{equation}
\theta_{(k)}^{\prime} \leftarrow \theta_{(k)}-\alpha \nabla_{\theta_{(k)}} \mathcal{L}(\mathcal{S}_{u}; \theta_{(k)}),
\label{method:innerloss}
\end{equation}
where $\mathcal{L}$ denotes the loss function mentioned in \equref{method:loss}, and $\alpha$ is the learning rate for adaptation. 
In the outer-loop optimization, we update $\theta$ on query set:
\begin{equation}
\theta_{(k+1)} \leftarrow \theta_{(k)}-\beta \nabla_{\theta_{(k)}} \mathcal{L}(\mathcal{Q}_{u}; \theta_{(k)}^{\prime}),
\label{method:metaloss}
\end{equation}
where $\beta$ is the learning rate. Note the loss in \equref{method:metaloss} is computed based on parameters $\theta_{(k)}^{\prime}$, which indicates the goal is to minimize model's estimation error on multiple cities' query set data after model's adaptation to a bit of support set data.
% such that the goal is to optimize model's parameters $\theta$ to fast adapt to all the tasks set.
% The meta-training procedure leverages the data of sampled cities $\{\mathcal{T}_{i}\}_{i=1}^{N_\mathcal{T}}$ to optimize $\theta$ alternately. 
After multiple of the above meta-training iterations in source cities are finished, the learned model's parameters containing extensive real estate market knowledge can be used as model initialization to fast adapt to a target city.

% However, the above meta-learning method only focus on the loss $\mathcal{L}$ over source cities, but overlook the impact of distribution difference between cities, which may introduce harmful source knowledge. 
% Motivated by CrossTReS~\cite{jin2022selective}, we design an adaptive instance re-weighting module to alleviate negative transfer. More formally, we rewrite Equation 12 as follows
However, the above meta-learning method overlooks the dynamic influence of source cities' instances for the target city with meta-training proceeding, which may introduce harmful source knowledge and lead to negative transfer.
\rev{
For example, during early meta-training, the model has limited knowledge of the target city and most source instances still provide helpful transferable patterns, \eg~general knowledge in real estate appraisal. As training proceeds, instances from source cities that exhibit divergent patterns to target distribution should be suppressed to maintain the effectiveness of knowledge transfer.}
To address the above problem, we introduce an instance re-weighting module to adaptively adjust the weights of source instances during meta-training. Formally, we rewrite \equref{method:metaloss} as below:
\begin{equation}
\theta_{(k+1)} \leftarrow \theta_{(k)} -\beta\sum_{s_{{n}} \in \mathcal{Q}_{u}} \lambda_{{n}} \nabla_{\theta_{(k)}} \mathcal{L}(s_{{n}}; \theta_{(k)}^{\prime}),
\label{method:outerloss}
\end{equation}
% \begin{equation}
% \mathcal{L}(s_t^*; \theta^{\prime})=(\hat{y}_{ij}-y_{ik})^2,
% \end{equation}
\rev{where $\lambda_{{n}}$ denotes learnable weight for each transaction instance $s_{{n}}$ in source cities, which aims to mitigate the influence of transaction instances that have negative effects on the target city, \eg~assigning smaller weights to source instances that exhibit patterns divergent from target city.}
We adopt a weight-generating network $\Omega(\cdot)$ to generate weight $\lambda_{{n}}$ for each transaction instance $s_{{n}}=\langle e_{{n}}, c_i, {{t}}, y_{{n}} \rangle$, defined as:
\begin{equation}
\lambda_{{n}}=\Omega \left([\mathbf{u} \mathbin\| \mathbf{z}^{{t}}_i \mathbin\| \mathbf{x}_{{n}} \mathbin\| y_{{n}}]\right),
\end{equation}
where $\mathbf{u}$ denotes identifier of the city that $s_{{n}}$ is from. 
In practice, we employ an MLP following a Sigmoid output function to instantiate $\Omega(\cdot)$, and denote the parameters of $\Omega(\cdot)$ as $\omega$.

\subsubsection{\textbf{Tri-Level Optimization}}
% \noindent{\textbf{Tri-Level Optimization.}}
Given model $\theta_{(k+1)}$ that have updated via \equref{method:outerloss},
the goal of instance weight $\lambda_{{n}}$ is that
after adapting model $\theta_{(k+1)}$ on the target city's training set, the model's estimation error on the target city's test set is minimum. 
To achieve this goal, we propose to upgrade the above bi-level optimization to a tri-level optimization based meta-learning framework.
% Denoting \equref{method:innerloss} as inner-loop and \equref{method:outerloss} as outer loop, 
In addition to the above inner-loop and outer-loop optimizations, we introduce an extra hyper-loop optimization to specifically update $\omega$ during meta-training by making full use of the target city's training set.
Specifically, we also split the target city's training set as a support set $\mathcal{S}_{tgt}$ and a query set $\mathcal{Q}_{tgt}$.
% Note that $\mathcal{S}^{target}$ is very small due to inherent data sparsity. 
Then, the hyper-loop optimization is formalized as:
% \vspace{-1mm}
\begin{equation}
\theta_{(k+1)}^{\prime \prime} \leftarrow \theta_{(k+1)} -\gamma_1 \nabla_{\theta_{(k+1)}} \mathcal{L}(\mathcal{S}_{tgt}; \theta_{(k+1)}),
\label{method:innerloss2}
\end{equation}
% \vspace{-2mm}
\begin{equation}
\omega_{(k+1)} \leftarrow \omega_{(k)}-\gamma_2 \nabla_{\lambda_{{n}}} \mathcal{L}(\mathcal{Q}_{tgt}; \theta_{(k+1)}^{\prime \prime}) \cdot \nabla_{\omega_{(k)}} \lambda_{{n}},
\end{equation}
where $\gamma_1$ and $\gamma_2$ are the learning rates. Through tri-level optimization, the weight-generating network is updated with the training proceeding to generate dynamic weights for source instances, which realizes adaptive knowledge distillation and mitigates negative knowledge transfer from source cities to the target city. 
Note all the learnable instance weights are generated by a shared weight-generating network that is independent of instances size, which guarantees the great scalability of our instance re-weighting module on large-scale data from the source cities.
\rev{
\subsection{Complexity Analysis}
In this section, we analyze the time complexity of MTTGN. Denote $|S|$ as the total number of potential transactions in a city to be appraised and $d$ as the dimension for all embeddings to ease the presentation. The time complexity of embedding update and embedding aggregation for all transactions would be $O(|S|d^2)$ and $O(|S||\mathcal{N}_i| d^2)$, respectively, where $|\mathcal{N}_i|$ is the number of neighboring communities of the transaction to be appraised. For HMTL module, the time complexity would be $O(|S|d^3)$. Hence, the overall computational time complexity for all potential transactions is $O(|S|(|\mathcal{N}_i|+d+1)d^2)$. In practice, the complexity of responding to an appraisal request would be $O((|\mathcal{N}_i|+d+1)d^2)$. Since the neighboring communities are limited within a certain radius, and the embedding dimension is set to relatively small, our model can be very efficient in the practical usage scenarios (see \secref{sec:latency}).
}

% \vspace{-3mm}
% -- Experiments
\section{Experiments}\label{sec:exp}
% In this section, we first describe the experimental setups, including data description, implementation details, evaluation metrics, and compared baselines. Then, we report and analyze the experimental results, encompassing the overall performance, ablation study, effect of different source city, effect of different transferred knowledge, analysis on instance weight, and efficiency test.
% This section analyzes the experimental results. More experiments for the analysis on instance weights and efficiency test are provided in \appref{app:weight} and \appref{sec:latency}.
% Implementation details of \meta are provided in \appref{app:implement}.

% \begin{table*}[tb]
%   % \small
%   \centering
%   \caption{Statistics of datasets.}
%   \vspace{-3mm}
%   \setlength{\tabcolsep}{2mm}{
%     \begin{tabular}{l|ccc|cc}
%       \toprule[1pt]
%       \multirow{2}{*}{\textbf{Description}} & \multicolumn{3}{c|}{Source Cities} & \multicolumn{2}{c}{Target Cities} \\
%       \cline{2-6} 
%       ~ & \chengdu & \wuhan & \guangzhou & \mianyang & \shaoxing \\
%       \midrule[0.5pt]
%       % Time scope & \multicolumn{3}{c|}{Jan 01, 2018 - Dec 31, 2019} & \multicolumn{2}{c}{Jan 01, 2018 - Dec 31, 2019}\\
%       \# of real estate transactions & 73,880 & 31,681 & 15,671 & 1,782 & 939 \\
%       \# of residential communities & 3,998 & 2,390 & 2,102 & 280 & 280 \\
%       \# of point of interests & 15,420 & 11,267 & 17,021 & 1,425 & 2,493 \\
%       \bottomrule[1pt]
%   \end{tabular}}
%   % \vspace{-3mm}
%   \label{table:dataset}
% \end{table*}

\begin{table}[tb]
  \small
  % \footnotesize
  % \vspace{-3mm}
  \centering
  \rev{
  \caption{Statistics of datasets.}
  \vspace{-3mm}
  \setlength{\tabcolsep}{2mm}{
    \begin{tabular}{l|ccc|ccc}
      \toprule[1pt]
      \multirow{2}{*}{\textbf{Description}} & \multicolumn{3}{c|}{Source Cities} & \multicolumn{3}{c}{Target Cities} \\
      \cline{2-7} 
      ~ & \chengdu & \wuhan & \guangzhou & \mianyang & \shaoxing & \zhuhai \\
      \midrule[0.5pt]
%       Time scope & \multicolumn{3}{c|}{Jan 01, 2018 - Dec 31, 2019} & \multicolumn{2}{c}{Jan 01, 2018 - Dec 31, 2019}\\
      \# of transactions & 73,880 & 31,681 & 15,671 & 1,782 & 939 & 894\\
      \# of communities & 3,998 & 2,390 & 2,102 & 280 & 280 & 278\\
      \# of POIs & 20,275 & 13,944 & 22,036 & 2,526 & 5,601 & 6,191\\
      % \# of POIs & 15,420 & 11,267 & 17,021 & 1,425 & 2,493 & xx\\
      \bottomrule[1pt]
  \end{tabular}}
  \label{table:dataset}
  }
  % \vspace{-3mm}
\end{table}

\begin{figure}[tb]
  \centering
%   \vspace{-1mm}
%   \includegraphics[width=0.98\columnwidth]{figs/statistics_of_transaction}
  \includegraphics[width=0.7\columnwidth]{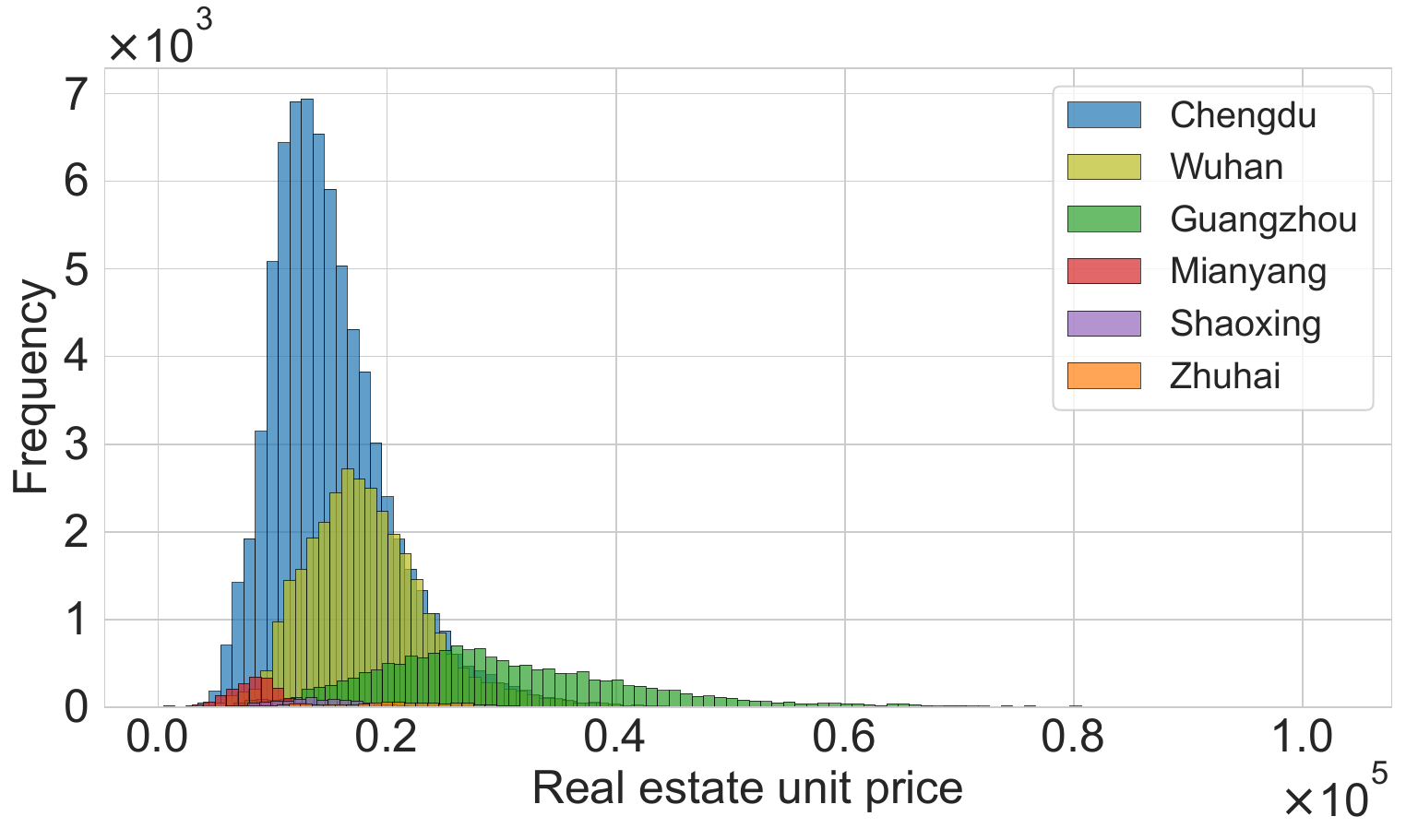}
  % \vspace{-2mm}
  \caption{\rev{Distributions of real estate unit price in different cities.}} 
  % \vspace{-3mm}
  \label{fig:district_price}
\end{figure}

\footnotetext[2]{\rev{The datasets are accessible in https://drive.google.com/drive/folders/1CHsEbbvfXdbTsSiY7hOszPkdVX1s2kNy}}
\footnotetext[3]{https://lianjia.com}

\subsection{Experimental setup}
% \label{sec:data_des}
\subsubsection{\textbf{Data Description}}\label{sec:data_des}
% \noindent \textbf{Data Description.}
\rev{We conduct experiments based on six real-world datasets\footnotemark[2] in correspondence to six representative cities in China, which are all collected from one of the world’s largest commercial real estate web platforms\footnotemark[3]. 
The statistics of each dataset are summarized in \tabref{table:dataset}.
The datasets include historical real estate transaction data, residential communities data, and Point of Interests~(POIs) data.
% The transaction record magnitudes for \chengdu, \wuhan, \guangzhou, \mianyang, \shaoxing are 73880, 31681, 15671, 1782, 939, respectively. The residential community magnitudes for these cities are 3998, 2390, 2102, 280, 280, and the POIs magnitudes are 15420, 11267, 17021, 1425, 2493, respectively. 
The used real estate apartment attributes mainly include the number of several kinds of rooms, house area, structure, decoration, orientation, heating method, floor number, whether tax free, transaction ownership, building type, elevator household ratio, completion year. The used community attributes primarily contain the number of subway stations, bus stations, kindergartens, primary and middle schools, colleges, hospitals,  pharmacies, malls, supermarkets, banks, restaurants, cafes, parks, cinemas, gymnasiums around the community, and the distance to the nearest above various facilities.
The real estate price distributions of six datasets are illustrated in \figref{fig:district_price}.

We choose three large cities, \chengdu, \wuhan, and \guangzhou, as source cities, and three small cities, \mianyang, \shaoxing, and \zhuhai, as target cities.
All source cities datasets range from January 01, 2018 to December 31, 2019. Target cities dataset \mianyang ranges from October 10, 2018 to December 31, 2019, \shaoxing ranges from September 17, 2018 to December 31, 2019, and \zhuhai ranges from January 02, 2018 to December 31, 2019, respectively.
We chronologically order each dataset based on the transaction time.
For each target city dataset, we respectively take the first 20, 100, and 500 transaction instances for training, and the rest for testing. 
For each source city dataset, we only use the data instances whose transaction times are not later than the target city's training instances. 
% Please refer to supplementary \secref{app:data} for the description of used features and illustration of datasets.
}

\subsubsection{\textbf{Implementation Details}}
\label{app:implement}
% \noindent \textbf{Implementation Details.}
All experiments are performed on a Linux server with 20-core Intel(R) Xeon(R) Platinum 8255C CPU @ 2.50GHz and NVIDIA Tesla V100 GPU.
We set $\epsilon=2000$ meters to connect neighboring communities. The dimension $d_{\phi}$ for time embedding is fixed to 8, and dimensions of $\mathbf{h}_i^{t}$ and $\widehat{\mathbf{h}}_i^{\hat{t}}$ are fixed to 64. 
% The hidden dimensions of $MLP$ is fixed to 64. 
The activation functions $\sigma$ in \equref{equ:weight} and \equref{equ:aggregate} are Tanh and ReLU, respectively. We use MLP to instantiate the hypernetworks. The MLPs implementation of hypernetworks, feature integration function, and weight-generating network are with ReLU activation and dimensions (16, 8), (64, 16), (64, 32) for hidden layers, respectively. 
We adopt Adam as meta-optimizer, set learning rates $\beta=0.01$, $\gamma_2=0.01$ to optimize model's parameters, and employ stochastic gradient descent for \equref{method:innerloss} and \equref{method:innerloss2} update by setting $\alpha=0.1$ and $\gamma_1=0.1$.
We train our model for 100 epochs on source cities during meta-training. 
For meta-testing on target city, we first determine a optimal adaptation step via splitting a validation set from the training set. Then we adapt our model for optimal adaptation steps on whole training set.

\subsubsection{\textbf{Evaluation Metrics}}
% \noindent \textbf{Evaluation Metrics.}
Three widely used metrics, including Mean Absolute Error~(MAE), 
Mean Absolute Percentage Error~(MAPE), and Root Mean Square Error~(RMSE), are adopted for performance evaluation.
% Note that for clear presentation of results, the estimated unit price and ground truth transaction unit price are both divided by 10,000 before inputting to these metrics.
The ground truth unit price and estimated unit price of real estate transactions are divided by 1,000 for clearer result presentation.

\subsubsection{\textbf{Baselines}}
% \noindent \textbf{Baselines.}
\rev{We compare \meta with eleven representative and competitive baselines, including one statistical baseline~(HA), two classic machine learning baselines~(LR~\cite{pedregosa2011scikit}, and GBRT~\cite{ke2017lightgbm}), three deep learning baselines~(DNN, MugRep~\cite{zhang2021mugrep}, and ST-RAP~\cite{lee2023st}), and five knowledge transfer baselines~(FT-DNN, FT-MTTGN, MAML-DNN~\cite{finn2017model}, MAML-MTTGN, and ST-GFSL~\cite{luCrossCityTransfer22}). 
Note that we do not select the models~\cite{wang2019cross,yao2019learning,jin2022selective,yuan2024spatio,wei2016transfer,ijcai2022p282} as baselines because they are specifically designed for regional spatio-temporal prediction tasks, whereas our problem does not incorporate the same region concept.}
% The details of these baselines are provided in \appref{app:baseline}.
% \subsection{Baseline Details}
% \label{app:baseline}
% \noindent \textbf{Baselines.}
% We compare \meta with eleven representative and competitive baselines, including six baselines without knowledge transfer~(HA, LR, GBRT, DNN, MugRep, ST-RAP), and six knowledge transfer baselines~(FT-DNN, FT-MTTGN, MAML-DNN, MAML-MTTGN, ST-GFSL). 
To ensure a fair comparison, we apply the same training strategy as \meta to all knowledge transfer-based baselines. 
As the baselines without knowledge transfer are only trained on the instances from the target city, we first determine an optimal training step by splitting a validation set with 20\% data from the training set, then we
re-train these models by the optimal training step on the whole training set.
We standardize the hidden dimensions to 64 and optimizer to Adam, carefully searching the key hyper-parameters of the baseline models around their recommended setups.
\begin{itemize}
\item~\textbf{HA}~uses the average price of historical transactions that occurred in the same residential community as estimated value. 
% For the case that no historical transactions in a community, HA uses average price of whole city's historical transactions. 
\item~\textbf{LR}~\cite{pedregosa2011scikit} appraises real estate through the well-known linear regression model, we use ridge regression implemented by scikit-learn.
% : \href{https://scikit-learn.org}{https://scikit-learn.org}.
\item~\textbf{GBRT}~\cite{ke2017lightgbm} appraises real estate through gradient boosted regression tree model. We set the maximal tree depth to 6, maximal leaves number to 31, and the learning rate to 0.1. We use the implementation by LightGBM.
% \href{https://lightgbm.readthedocs.io/}{https://lightgbm.-readthedocs.io/}.
\item~\textbf{DNN}~appraises real estate through deep neural networks, 
which have been widely used in real estate appraisal~\cite{peter2020review}. We set the learning rate to 0.01.
We implement it via three fully connected layers with ReLU activation functions. 
\item~\textbf{MugRep}~\cite{zhang2021mugrep} is a state-of-the-art GNN-based approach for real estate appraisal. It employs a hierarchical graph representation learning framework to model the correlations among transactions and communities. We set the learning rate to 0.01, and the number of GNN layers to 1 for both event-level and community-level representation learning.
\item~\textbf{ST-RAP}~\cite{lee2023st} is a state-of-the-art GNN-based approach for real estate appraisal. It employs heterogeneous graph neural networks to capture temporal dynamics and spatial relationships among real estate transactions simultaneously. We set the learning rate to 0.01.
\item~\textbf{FT-DNN}~is trained on mixed datasets of all source cities and fine-tuned on the target city and uses DNN as the base model. 
% Note we can not compare the transfer learning version of MugRep as its learned district-specific valuation parameters are dedicated to a specific city without generalization ability across cities. 
We set the learning rate to 0.01 for pre-training and 0.1 for fine-tuning.
\item~\textbf{FT-MTTGN}~is similar to FT-DNN but uses our proposed MTTGN as the base model. We set the learning rate to 0.01 for pre-training and 0.1 for fine-tuning.
\item~\textbf{MAML-DNN}~\cite{finn2017model} is a powerful model-agnostic meta-learning method, which aims to learn a good initialization from multiple source cities for fast adaptation to the target city, and uses DNN as the base model. We set the learning rate to 0.01 for meta-training and 0.1 for meta-testing.
\item~\textbf{MAML-MTTGN} is similar to MAML-DNN but uses our proposed MTTGN as the base model. We set the learning rate to 0.01 for meta-training and 0.1 for meta-testing.
\item~\textbf{ST-GFSL}~\cite{luCrossCityTransfer22} is a state-of-the-art cross-city knowledge transfer method for graph-based traffic prediction tasks, which adopts MAML to learn spatiotemporal meta-knowledge across cities to generate a GRU-based model's parameters. We adapt ST-GFSL to our task by regarding each community as a node. We set the dimension of meta-knowledge to 16, the learning rate to 0.01 for meta-training and 0.1 for meta-testing.
\end{itemize}

\rev{
\begin{table*}[tb]
  % \small
  \scriptsize
  % \footnotesize
  % \vspace{-3mm}
  \centering
  \rev{
  \caption{\rev{Overall performance evaluated by MAE, MAPE, RMSE on three target cities with different number of training instances. \textbf{Bold} represents the best-performing results, and \underline{underline} denotes the second-best results. The improvement of \meta over the baselines is statistically significant under the Wilcoxon signed-rank test~(p < 0.05).}
  }
  \vspace{-3mm}
%   \resizebox{\wfill}{!}
  \setlength{\tabcolsep}{0.8mm}{
    \begin{tabular}{c|c|ccc|ccc|ccc}
      \toprule[1pt]
        Data & \multirow{2}{*}{Algorithm} & \multicolumn{3}{c|}{20 Training Instances } & \multicolumn{3}{c|}{100 Training Instances} & \multicolumn{3}{c}{500 Training Instances}\\
        \cline{3-11} 
        set & ~ & MAE & MAPE & RMSE & MAE & MAPE & RMSE & MAE & MAPE & RMSE \\
      \midrule[0.5pt]
      \multirow{12}{*}{\rotatebox{90}\mianyang} & HA & 1.709 & 24.89\% & 2.211 & 1.489 & 22.27\% & 1.971 & 1.205 & 17.16\% & 1.685\\
      ~ & LR & 1.612 & 23.63\% & 2.085 & 1.394 & 20.16\% & 1.831 & 1.061 & 14.22\% & 1.412\\
      ~ & GBRT & 1.719 & 25.10\% & 2.222 & 1.370 & 19.63\% & 1.808 & 1.045 & 13.77\% & 1.409\\
      ~ & DNN & \valstd{1.679}{0.03} & \valstd{23.71\%}{0.3} & \valstd{2.162}{0.03} & \valstd{1.408}{0.07} & \valstd{19.20\%}{1.3} & \valstd{1.845}{0.08} & \valstd{1.115}{0.02} & \valstd{14.56\%}{0.4} & \valstd{1.506}{0.02}\\ 
      ~ & MugRep & \valstd{1.771}{0.04}	& \valstd{25.68\%}{0.7} & \valstd{2.422}{0.35} & \valstd{1.451}{0.09}	& \valstd{21.06\%}{1.3}	& \valstd{1.951}{0.12} & \valstd{0.982}{0.04}	& \valstd{13.05\%}{0.7} & \valstd{1.340}{0.05} \\
      ~ & ST-RAP & \valstd{1.535}{0.08}	& \valstd{21.86\%}{1.0} & \valstd{2.011}{0.09} & \valstd{1.449}{0.05} & \valstd{20.48\%}{0.9} & \valstd{1.885}{0.05} & \valstd{1.068}{0.04} & \valstd{14.60\%}{0.7} & \valstd{1.445}{0.06} \\
      % ~ & name & x	& x\% & x & x & x\%	& x & x	& x\% & x \\
      ~ & FT-DNN & \valstd{1.425}{0.02} & \valstd{19.76\%}{0.2} & \valstd{1.869}{0.02} & \valstd{1.266}{0.02} & \valstd{17.61\%}{0.2} & \valstd{1.685}{0.02} & \valstd{1.036}{0.01} & \valstd{13.68\%}{0.3} & \valstd{1.414}{0.01}\\
      ~ & FT-MTTGN & \valstd{1.159}{0.01} & \valstd{16.01\%}{0.2} & \valstd{1.588}{0.02} & \valstd{1.109}{0.01} & \valstd{15.42\%}{0.2} & \valstd{1.493}{0.02} & \valstd{0.913}{0.01} & \valstd{11.55\%}{0.2} & \valstd{1.258}{0.01}\\
      ~ & MAML-DNN & \valstd{1.342}{0.03} & \valstd{18.85\%}{0.5} & \valstd{1.763}{0.04} & \valstd{1.241}{0.03} & \valstd{16.96\%}{0.4} & \valstd{1.645}{0.03} & \valstd{1.018}{0.01} & \valstd{13.56\%}{0.3} & \valstd{1.402}{0.03}\\
      ~ & MAML-MTTGN & \valstdu{1.102}{0.04} & \valstdu{15.83\%}{0.6} & \valstdu{1.492}{0.06} & \valstdu{1.056}{0.02} & \valstdu{14.47\%}{0.4} & \valstdu{1.458}{0.02} & \valstdu{0.866}{0.01} & \valstdu{11.01\%}{0.2} & \valstdu{1.208}{0.01} \\
      ~ & ST-GFSL & \valstd{1.202}{0.06} & \valstd{17.24\%}{1.0} & \valstd{1.601}{0.06} & \valstd{1.161}{0.03} & \valstd{15.99\%}{0.4} & \valstd{1.561}{0.03} & \valstd{0.933}{0.01} & \valstd{12.12\%}{0.1} & \valstd{1.299}{0.02} \\
      ~ & \meta & \valstdb{0.996}{0.01} & \valstdb{13.21\%}{0.1} & \valstdb{1.386}{0.01} & \valstdb{0.954}{0.02} & \valstdb{12.97\%}{0.4} & \valstdb{1.318}{0.03} & \valstdb{0.836}{0.03} & \valstdb{10.61\%}{0.5} & \valstdb{1.164}{0.04} \\
      \midrule[0.5pt]
      \multirow{12}{*}{\rotatebox{90}\shaoxing} & HA & 3.223 & 22.58\% & 4.425 & 2.960 & 21.49\% & 4.168 & 2.826 & 19.90\% & 4.231\\
      ~ & LR & 3.102 & 22.40\% & 4.246 & 2.800 & 20.62\% & 3.896 & 2.643 & 18.02\% & 3.996\\
      ~ & GBRT & 3.364 & 23.99\% & 4.482 & 2.904 & 20.97\% & 4.012 & 2.524 & 17.29\% & 3.751\\
      ~ & DNN & \valstd{3.286}{0.09} & \valstd{23.75\%}{0.3} & \valstd{4.393}{0.10} & \valstd{2.941}{0.06} & \valstd{20.23\%}{0.3} & \valstd{4.005}{0.07} & \valstd{2.636}{0.05} & \valstd{18.18\%}{0.6} & \valstd{3.954}{0.05}\\
      ~ & MugRep & \valstd{3.316}{0.04} & \valstd{23.22\%}{0.3} & \valstd{4.694}{0.09} & \valstd{3.005}{0.09} & \valstd{22.23\%}{0.4} & \valstd{4.219}{0.41} & \valstd{2.682}{0.09} & \valstd{17.99\%}{0.9} & \valstd{4.085}{0.08}\\
      ~ & ST-RAP & \valstd{3.471}{0.19} & \valstd{24.42\%}{1.2} & \valstd{4.58}{0.20} & \valstd{3.065}{0.05} & \valstd{22.28\%}{0.4} & \valstd{4.239}{0.08} & \valstd{2.675}{0.05} & \valstd{17.71\%}{0.5} & \valstd{4.066}{0.04} \\
      % ~ & name & x	& x\% & x & x & x\%	& x & x	& x\% & x \\
      ~ & FT-DNN & \valstd{2.841}{0.04} & \valstd{20.95\%}{0.3} & \valstd{3.948}{0.06} & \valstd{2.673}{0.04} & \valstd{18.95\%}{0.5} & \valstd{3.780}{0.05} & \valstd{2.416}{0.03} & \valstd{16.40\%}{0.3} & \valstd{3.714}{0.05}\\
      ~ & FT-MTTGN & \valstd{2.433}{0.03} & \valstd{17.15\%}{0.4} & \valstd{3.441}{0.04} & \valstd{2.361}{0.02} & \valstd{16.78\%}{0.1} & \valstd{3.436}{0.04} & \valstd{2.201}{0.06} & \valstd{15.02\%}{0.3} & \valstd{3.507}{0.10} \\
      ~ & MAML-DNN & \valstd{2.694}{0.03} & \valstd{18.73\%}{0.2} & \valstd{3.811}{0.08} & \valstd{2.542}{0.02} & \valstd{18.21\%}{0.6} & \valstd{3.548}{0.04} & \valstd{2.332}{0.02} & \valstd{16.02\%}{0.3} & \valstd{3.660}{0.04}\\
      ~ & MAML-MTTGN & \valstdu{2.362}{0.02} & \valstdu{17.04\%}{0.2} & \valstdu{3.435}{0.05} & \valstdu{2.232}{0.02} & \valstdu{15.91\%}{0.2} & \valstdu{3.318}{0.03} & \valstdu{2.150}{0.04} & \valstdu{14.71\%}{0.4} & \valstdu{3.362}{0.06} \\
      ~ & ST-GFSL & \valstd{2.512}{0.03} & \valstd{18.09\%}{0.2} & \valstd{3.710}{0.05} & \valstd{2.372}{0.03} & \valstd{17.02\%}{0.3} & \valstd{3.483}{0.07} & \valstd{2.262}{0.03} & \valstd{15.41\%}{0.2} & \valstd{3.572}{0.03}\\
      ~ & \meta & \valstdb{2.205}{0.02} & \valstdb{16.09\%}{0.2} & \valstdb{3.228}{0.04} & \valstdb{2.080}{0.04} & \valstdb{14.95\%}{0.4} & \valstdb{3.132}{0.03} & \valstdb{1.982}{0.03} & \valstdb{13.78\%}{0.2} & \valstdb{3.058}{0.05} \\
      \midrule[0.5pt]
      \multirow{12}{*}{\rotatebox{90}\zhuhai} & HA & 7.531 & 54.16\% & 9.313 & 5.038 & 30.76\% & 6.901 & 2.771 & 15.90\% & 4.408\\
      ~ & LR & 7.261 & 52.10\% & 9.058 & 4.669 & 24.64\% & 6.430 & 2.631 & 14.97\% & 3.937\\
      ~ & GBRT & 7.987 & 57.64\% & 9.745 & 5.100 & 30.57\% & 6.491 & 2.667 & 14.21\% & 3.898\\
      ~ & DNN & \valstd{6.409}{0.20} & \valstd{45.10\%}{1.6} & \valstd{8.119}{0.21} & \valstd{5.088}{0.24} & \valstd{27.06\%}{1.1} & \valstd{6.756}{0.27} & \valstd{2.963}{0.20} & \valstd{14.58\%}{1.0} & \valstd{4.427}{0.21}\\
      ~ & MugRep & \valstd{6.950}{0.28} & \valstd{46.80\%}{1.9} & \valstd{8.810}{0.29} & \valstd{5.890}{0.25} & \valstd{29.50\%}{1.5} & \valstd{7.560}{0.25} & \valstd{3.030}{0.23} & \valstd{15.60\%}{1.4} & \valstd{4.520}{0.24}\\
      ~ & ST-RAP & \valstd{6.690}{0.15} & \valstd{45.80\%}{0.9} & \valstd{8.591}{0.19} & \valstd{6.218}{0.11} & \valstd{35.42\%}{0.6} & \valstd{7.878}{0.20} & \valstd{3.938}{0.18} & \valstd{19.61\%}{0.8} & \valstd{5.381}{0.23} \\
      % ~ & name & x	& x\% & x & x & x\%	& x & x	& x\% & x \\
      ~ & FT-DNN & \valstd{5.327}{0.05} & \valstd{33.79\%}{0.4} & \valstd{7.028}{0.08} & \valstd{4.578}{0.03} & \valstd{24.87\%}{0.4} & \valstd{6.290}{0.13} & \valstd{3.201}{0.12} & \valstd{16.31\%}{0.5} & \valstd{4.585}{0.19}\\
      ~ & FT-MTTGN & \valstd{3.899}{0.24} & \valstd{21.72\%}{1.4} & \valstd{5.373}{0.18} & \valstd{3.271}{0.10} & \valstd{18.36\%}{0.5} & \valstd{4.716}{0.10} & \valstd{2.468}{0.06} & \valstd{13.09\%}{0.5} & \valstd{3.752}{0.07} \\
      ~ & MAML-DNN & \valstd{5.083}{0.03} & \valstd{30.25\%}{0.6} & \valstd{6.823}{0.10} & \valstd{4.341}{0.05} & \valstd{21.95\%}{0.6} & \valstd{6.081}{0.12} & \valstd{3.069}{0.04} & \valstd{15.80\%}{0.4} & \valstd{4.475}{0.12}\\
      ~ & MAML-MTTGN & \valstdu{3.486}{0.12} & \valstdu{20.00\%}{0.9} & \valstdu{5.024}{0.15} & \valstdu{3.007}{0.08} & \valstdu{15.66\%}{0.4} & \valstdu{4.518}{0.07} & \valstdu{2.331}{0.05} & \valstdu{12.01\%}{0.3} & \valstdu{3.669}{0.10} \\
      ~ & ST-GFSL & \valstd{4.303}{0.14} & \valstd{26.13\%}{0.9} & \valstd{5.798}{0.12} & \valstd{3.467}{0.11} & \valstd{18.20\%}{0.3} & \valstd{5.145}{0.17} & \valstd{2.692}{0.05} & \valstd{13.28\%}{0.2} & \valstd{4.229}{0.06}\\
      ~ & \meta & \valstdb{3.237}{0.08} & \valstdb{18.28\%}{0.6} & \valstdb{4.739}{0.11} & \valstdb{2.924}{0.08} & \valstdb{14.71\%}{0.6} & \valstdb{4.337}{0.10} & \valstdb{2.225}{0.10} & \valstdb{10.53\%}{0.5} & \valstdb{3.553}{0.19} \\
      \bottomrule[1pt]
  \end{tabular}}
  % \vspace{-3mm}
  \label{table:overall_results}
  }
\end{table*}  
}

\subsection{Overall Performance}
\rev{\tabref{table:overall_results} report the performance of our model and all compared baselines on three target city datasets with different number of training instances in target city \wrt three metrics. To mitigate randomness, we run each experiment using five different random seeds and report the mean and standard deviation of the results.
Consistently, \meta achieves the best performance across all datasets with different training instances and demonstrates a statistically significant improvement over baselines under the Wilcoxon signed-rank test.
Moreover, we have several observations. 
Firstly, the deep learning approaches with knowledge transfer~(\ie FT-DNN, FT-MTTGN, MAML-DNN, MAML-MTTGN, ST-GFSL, and \meta) basically outperform the approaches without knowledge transfer~(\ie HA, LR, GBRT, DNN, MugRep, ST-RAP), which verifies that extensive valuable knowledge can be transferred from data-rich sources cities to substantially improve deep learning model’s performance in data-scarce target city. Especially, the improvement is most pronounced when target cities only have very few 20 training instances. 
Secondly, by comparing meta-learning approaches to fine-tuning approaches with the same base models~(\ie MAML-DNN vs FT-DNN, MAML-MTTGN vs FT-MTTGN),
we observe meta-learning approaches have superior performance. 
This is because MAML is designed for learning generalizable knowledge from a set of tasks drawn from multiple source cities, enabling rapid adaptation to a new task in the target city—an approach well suited for cross-city real estate appraisal with multi-source data. By contrast, fine-tuning approaches simply mix multi-source data into one dataset, ignoring the heterogeneity across source cities. 
Thirdly, we find knowledge transfer approaches with MTTGN as base model~(\ie FT-MTTGN, MAML-MTTGN, \meta) surpass approaches FT-DNN, MAML-DNN, ST-GFSL, demonstrating the more powerful capability of MTTGN to capture the complicated association factors, \eg~irregular spatiotemporal correlation and price diversity among real estate transactions, for accurate real estate appraisal.
Lastly, we observe \meta are superior to the best baseline MAML-MTTGN though they are with the same base model MTTGN. The reason is that \meta introduces instances adaptive re-weighting with tri-level optimization framework, which effectively mitigates negative knowledge transfer from the instance level during meta-training. 
}

\begin{figure}[tb]
  % \vspace{-4mm}
  \centering
  \hspace{-1mm}
  \subfigure[{MAE}]{
    \includegraphics[width=0.335\columnwidth]{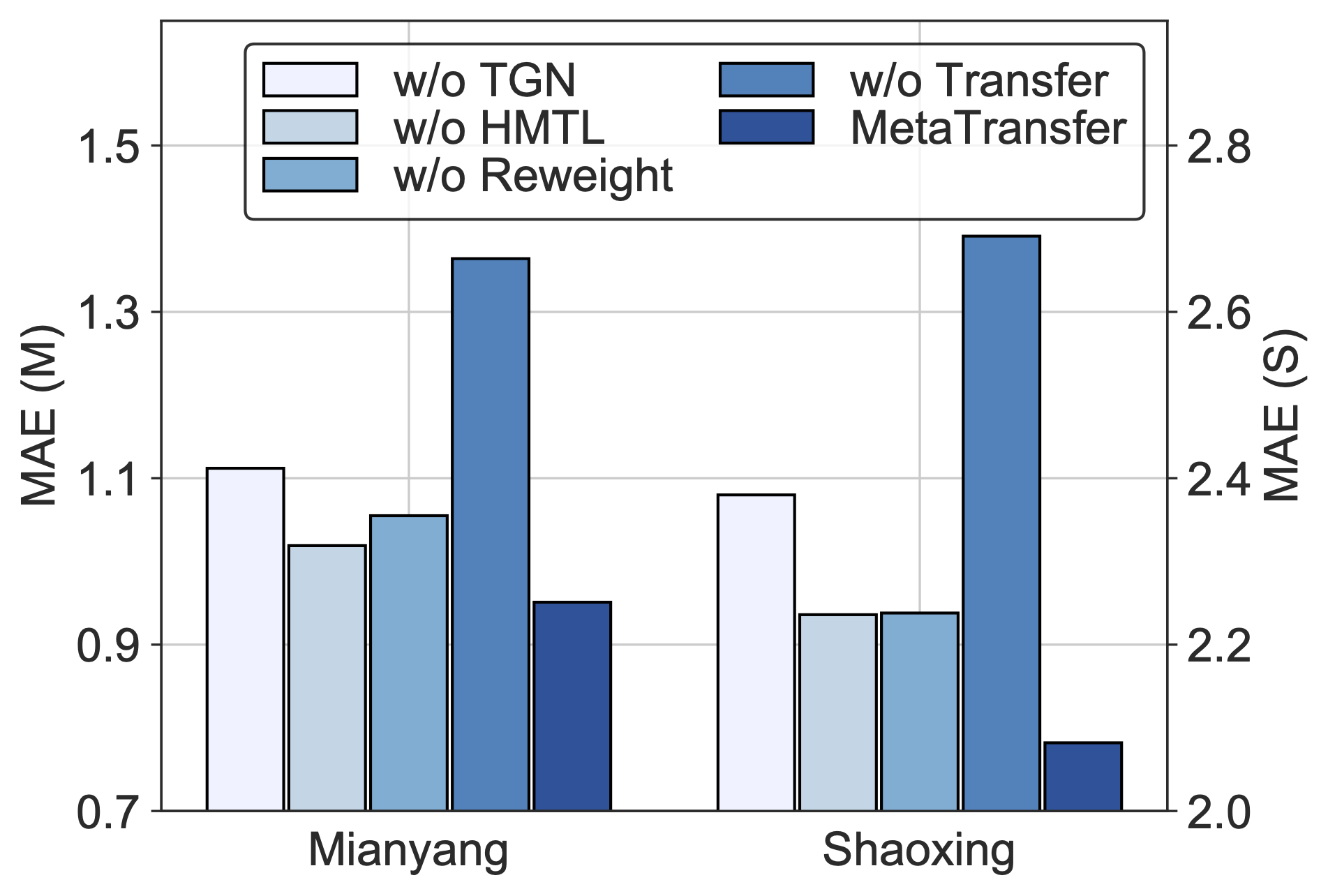}}\hspace{-1mm}
  \subfigure[{MAPE}]{
    \includegraphics[width=0.3\columnwidth]{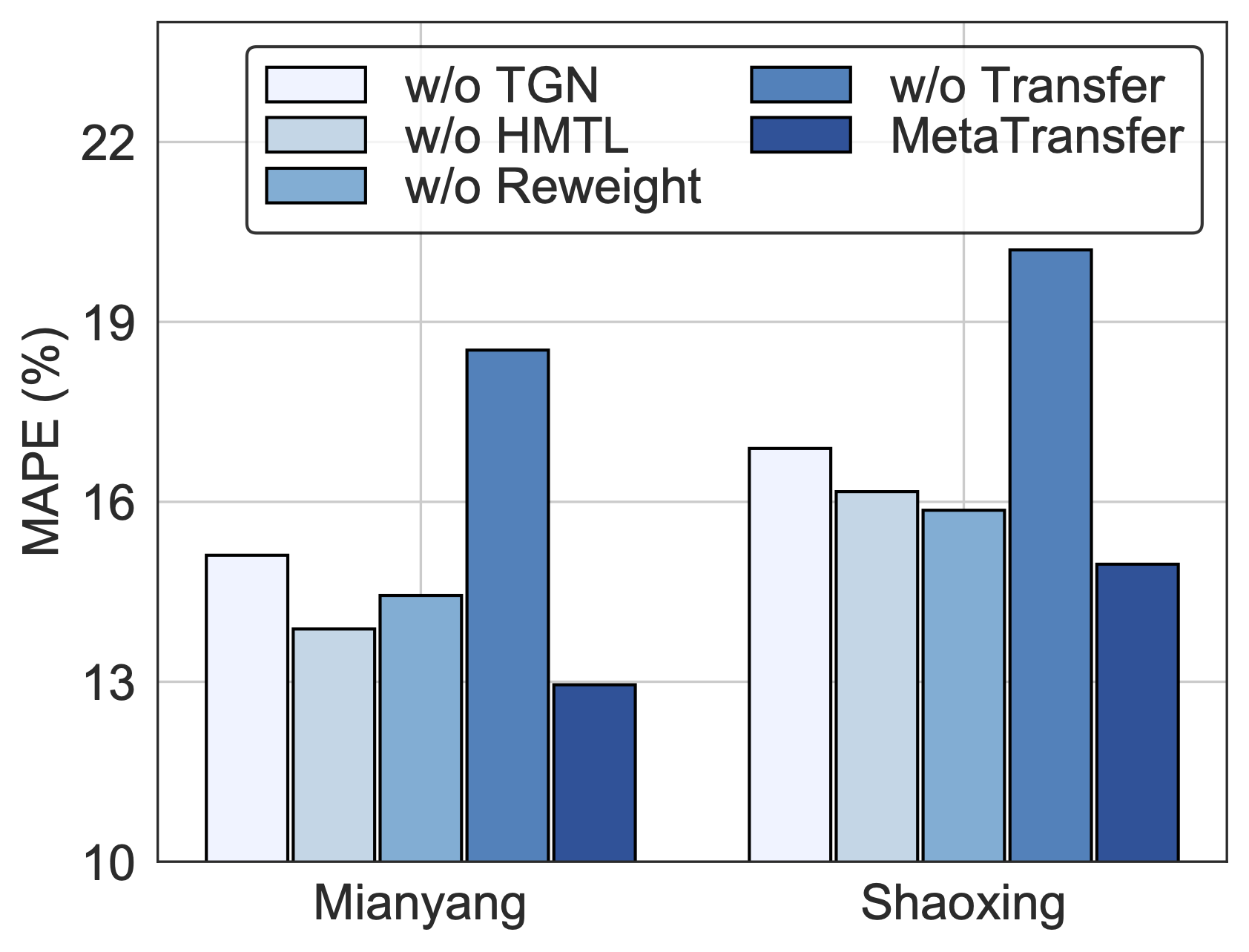}}\hspace{-0.5mm}
  \subfigure[{RMSE}]{
    \includegraphics[width=0.34\columnwidth]{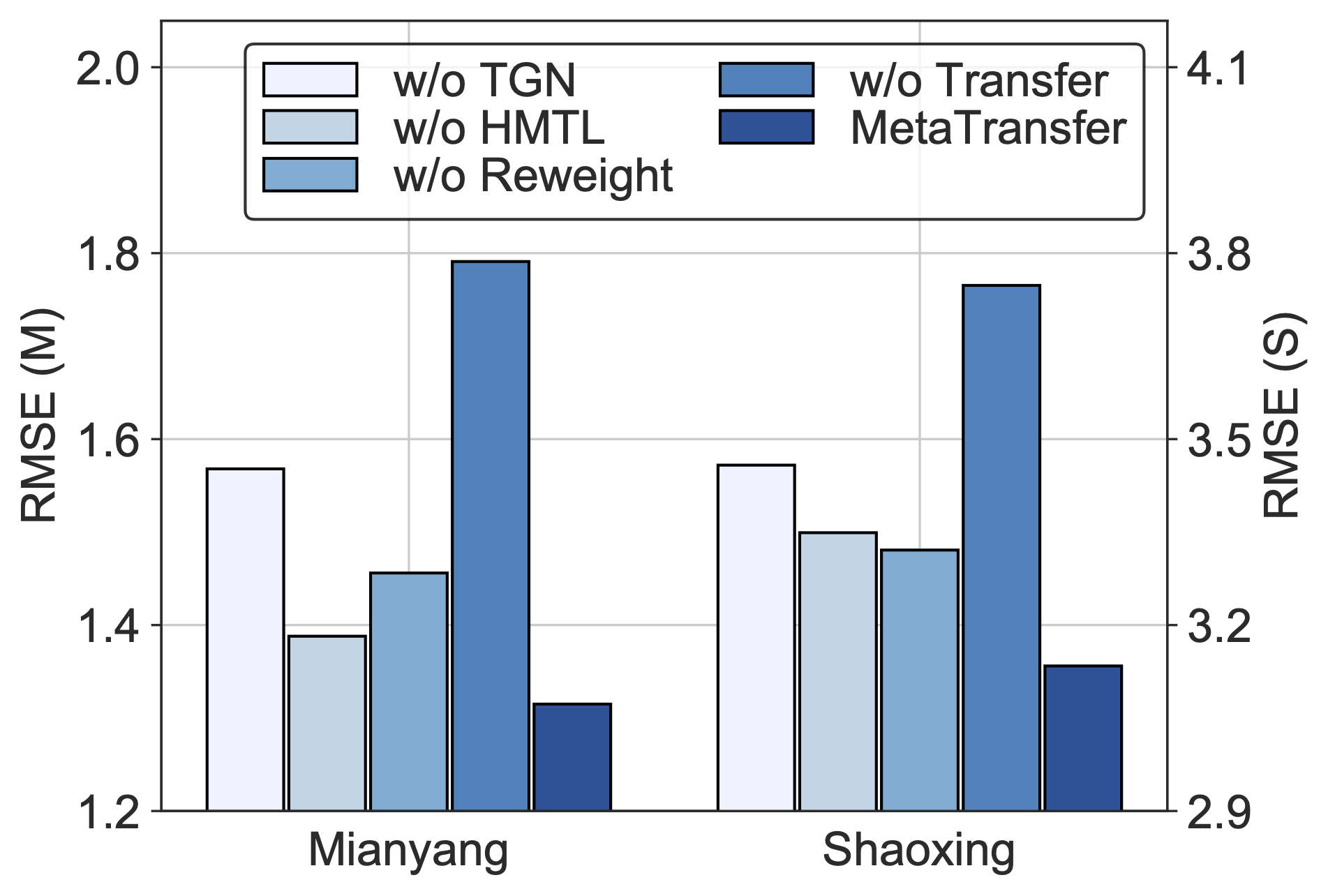}}
  \vspace{-3mm}
  \caption{Ablation study. ``M'' and ``S'' denote \mianyang and \shaoxing, respectively.} 
  % \vspace{-5mm}
  \label{fig:ablation_study}
\end{figure}

\subsection{Ablation Study}
We evaluate the performance of \meta and its four variants on \mianyang and \shaoxing with 100 training instances for three metrics:
(1)~\textbf{w/o TGN} removes the TGN module;
(2)~\textbf{w/o HMTL} removes the HMTL module, thus all communities have the shared parameters;
(3)~\textbf{w/o Reweight} removes instance re-weighting and tri-level optimization;
(4)~\textbf{w/o Transfer} trains base model on target city without cross-city knowledge transfer. 
As can be seen in \figref{fig:ablation_study}, removing any component leads to remarkable performance degradation. 
Particularly, the model without cross-city knowledge transfer causes a extremely performance degradation on all datasets, which demonstrates the effectiveness to improve real estate appraisal in data-scarce cities through meta-learning.
In addition, removing TGN results in significant performance degradation, which verifies TGN's effect on modeling irregular spatiotemporal correlations of real estate transactions.
We further observe the model without instance re-weighting also leads to an obvious performance decline on all the metrics and datasets, this is because there exists negative knowledge transfer from some instances of source cities to target city, while instance re-weighting with tri-level optimization can mitigate this problem.
By comparing \meta with \textbf{w/o HMTL}, it indicates effectiveness of HMTL module to simultaneously facilitate intra-city knowledge sharing and distinguish real estate price distribution between communities.

\begin{figure}[tb]
 % \vspace{-2mm}
  \centering
  \includegraphics[width=0.8\columnwidth]{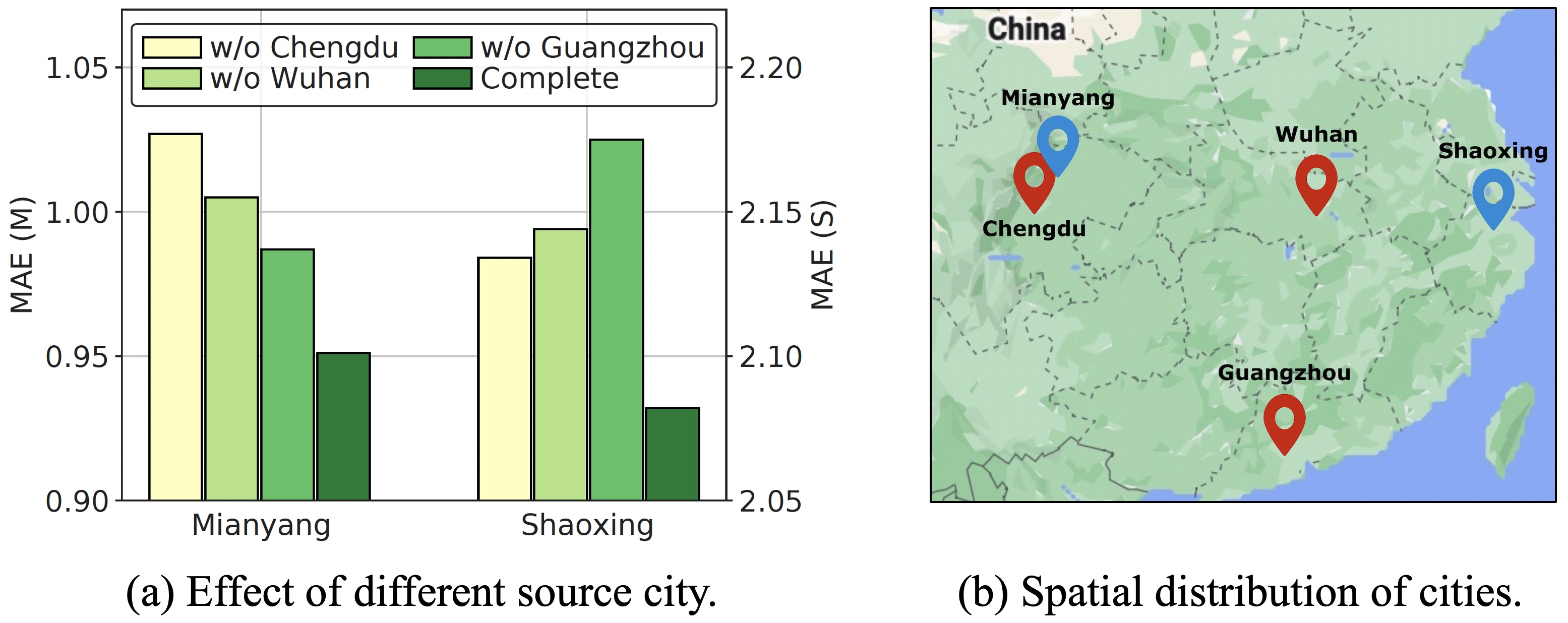}
  % \vspace{-5mm}
  \caption{Effect of different source cities on cross-city knowledge transfer.} 
  \vspace{-3mm}
  \label{fig:city_effect}
\end{figure}

% \begin{figure}[tb]
%   \vspace{-2mm}
%   \centering
%   % \vspace{-2mm}
%   \subfigure[{MAE.}]{
%     \includegraphics[width=0.325\columnwidth]{figs/feature_MAE.png}}\hspace{-1mm}
%   \subfigure[{MAPE.}]{
%     \includegraphics[width=0.285\columnwidth]{figs/feature_MAPE.png}}\hspace{-0.5mm}
%   \subfigure[{RMSE.}]{
%     \includegraphics[width=0.325\columnwidth]{figs/feature_RMSE.png}}
%   % \vspace{-5mm}
%   \caption{Effect of transferred knowledge from a features perspective.} 
%   % \vspace{-6mm}
%   \label{fig:feature_perspective}
% \end{figure}

% \begin{figure}[tb]
%     % \vspace{-3mm}
%   \centering
% %   \hspace{-1mm}
% %   \vspace{-1mm}
%   \subfigure[{MAE.}]{
%     \includegraphics[width=0.325\columnwidth]{figs/parameter_MAE.png}}\hspace{-1mm}
%   \subfigure[{MAPE.}]{
%     \includegraphics[width=0.285\columnwidth]{figs/parameter_MAPE.png}}\hspace{-0.5mm}
%   \subfigure[{RMSE.}]{
%     \includegraphics[width=0.325\columnwidth]{figs/parameter_RMSE.png}}
%   % \vspace{-5mm}
%   \caption{Effect of transferred knowledge from a parameters perspective.} 
%   \vspace{-3mm}
%   \label{fig:parameter_perspective}
% \end{figure}

\begin{figure}[tb]
  \centering
  % \vspace{-2mm}
  \subfigure[{MAE}]{
    \includegraphics[width=0.335\columnwidth]{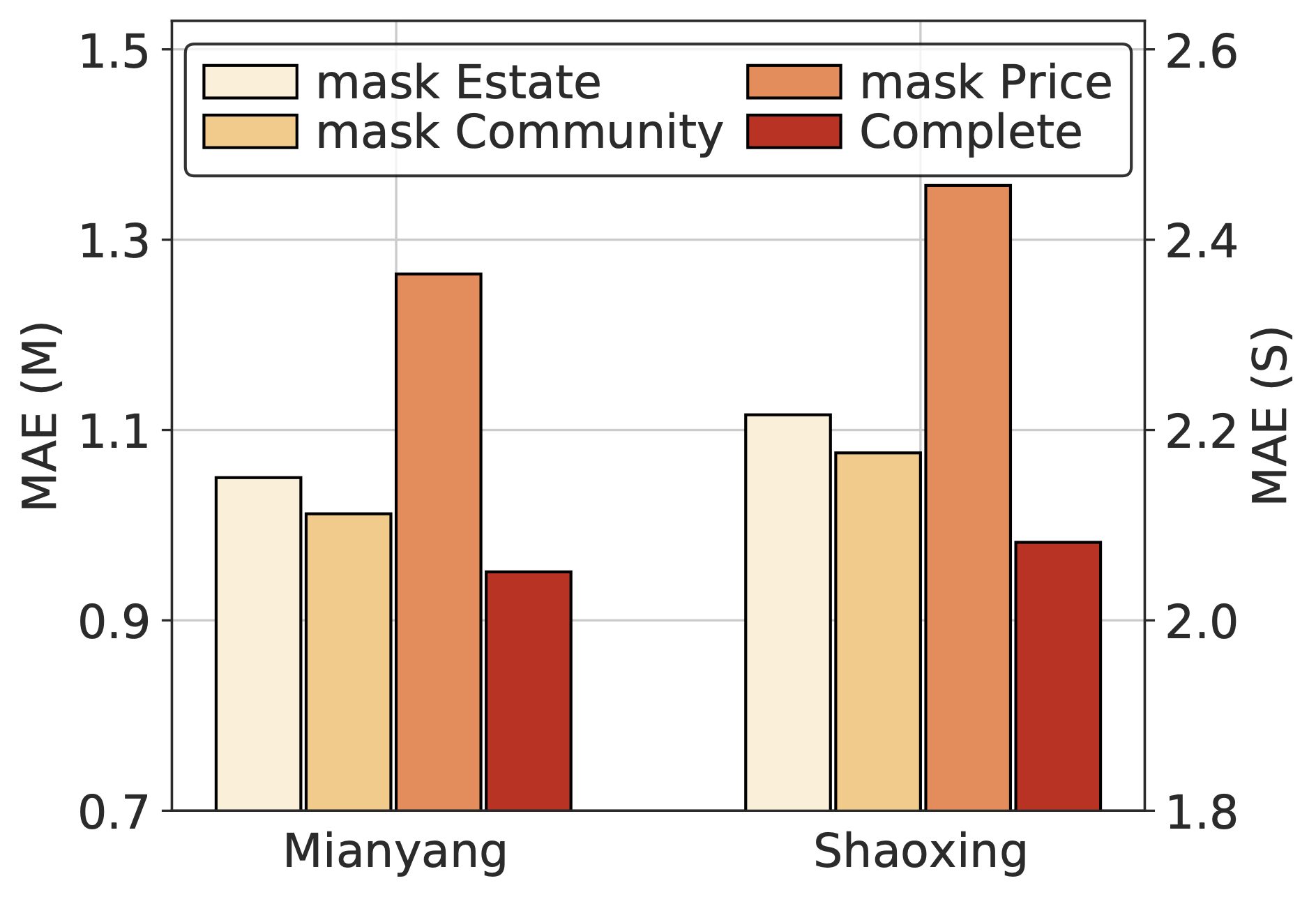}}\hspace{-1mm}
  \subfigure[{MAPE}]{
    \includegraphics[width=0.295\columnwidth]{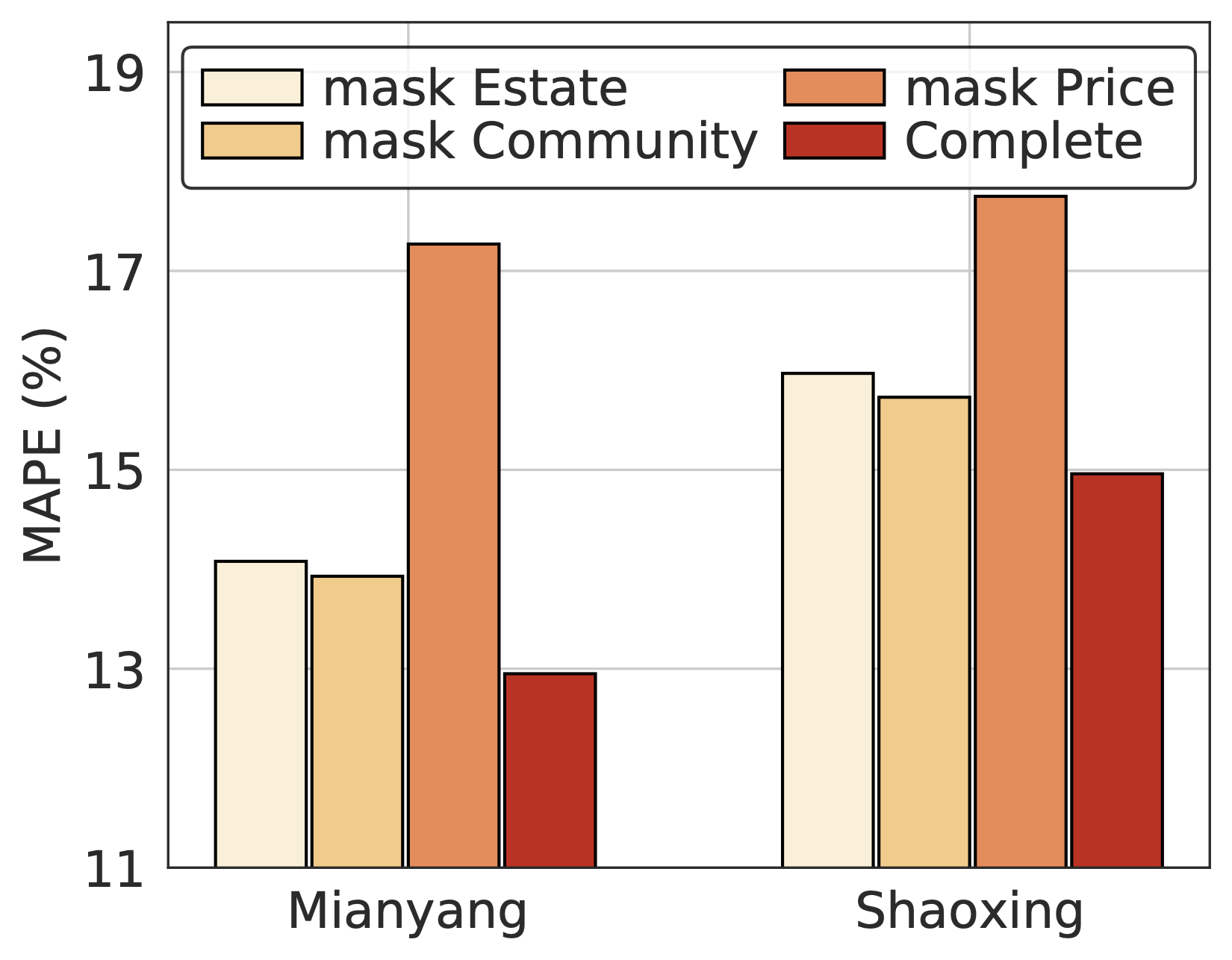}}\hspace{-0.5mm}
  \subfigure[{RMSE}]{
    \includegraphics[width=0.34\columnwidth]{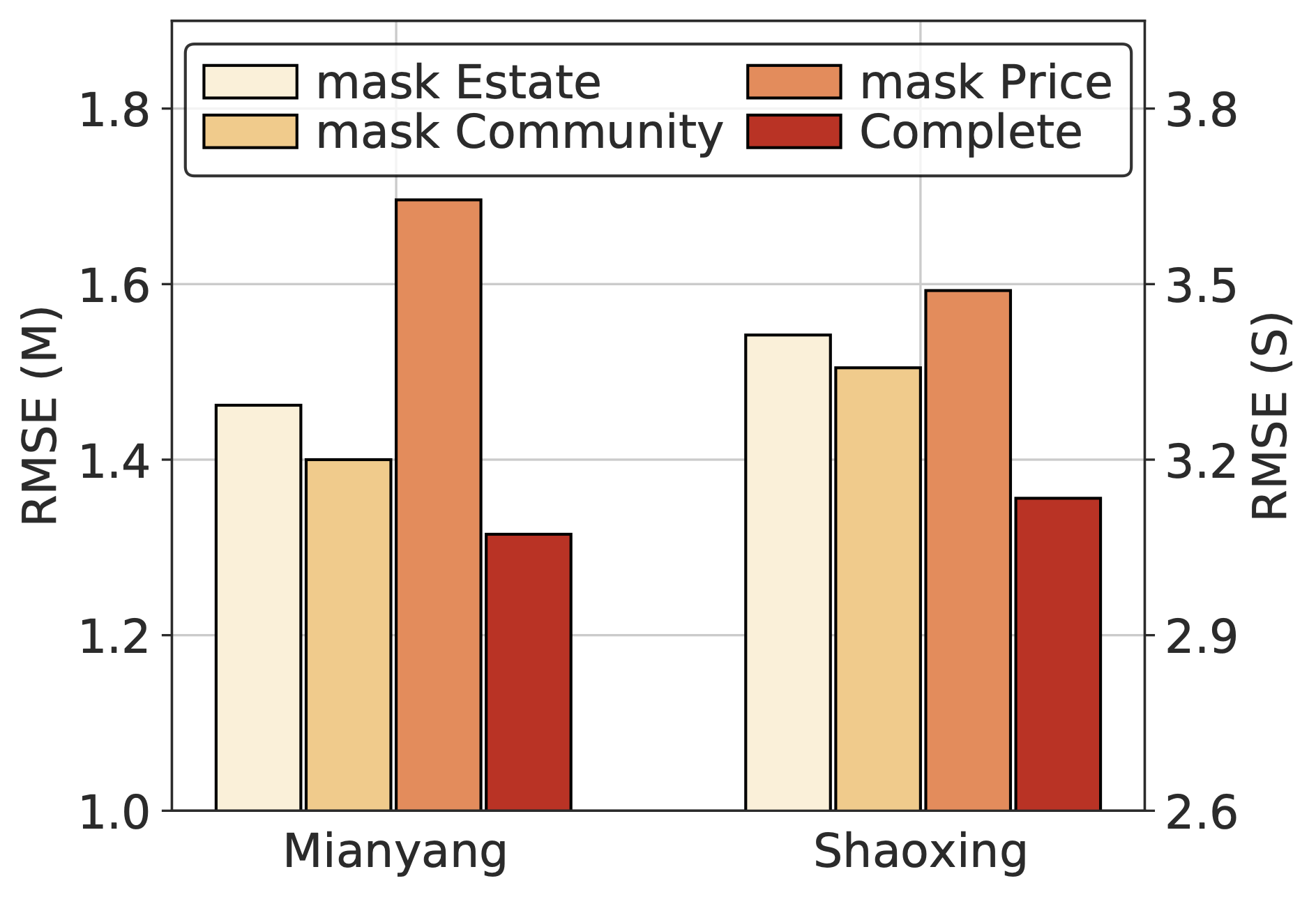}}
  \vspace{-3mm}
  \caption{Effect of transferred knowledge from a features perspective.} 
  \vspace{-3mm}
  \label{fig:feature_perspective}
\end{figure}

\begin{figure}[tb]
    % \vspace{-3mm}
  \centering
%   \hspace{-1mm}
%   \vspace{-1mm}
  \subfigure[{MAE}]{
    \includegraphics[width=0.335\columnwidth]{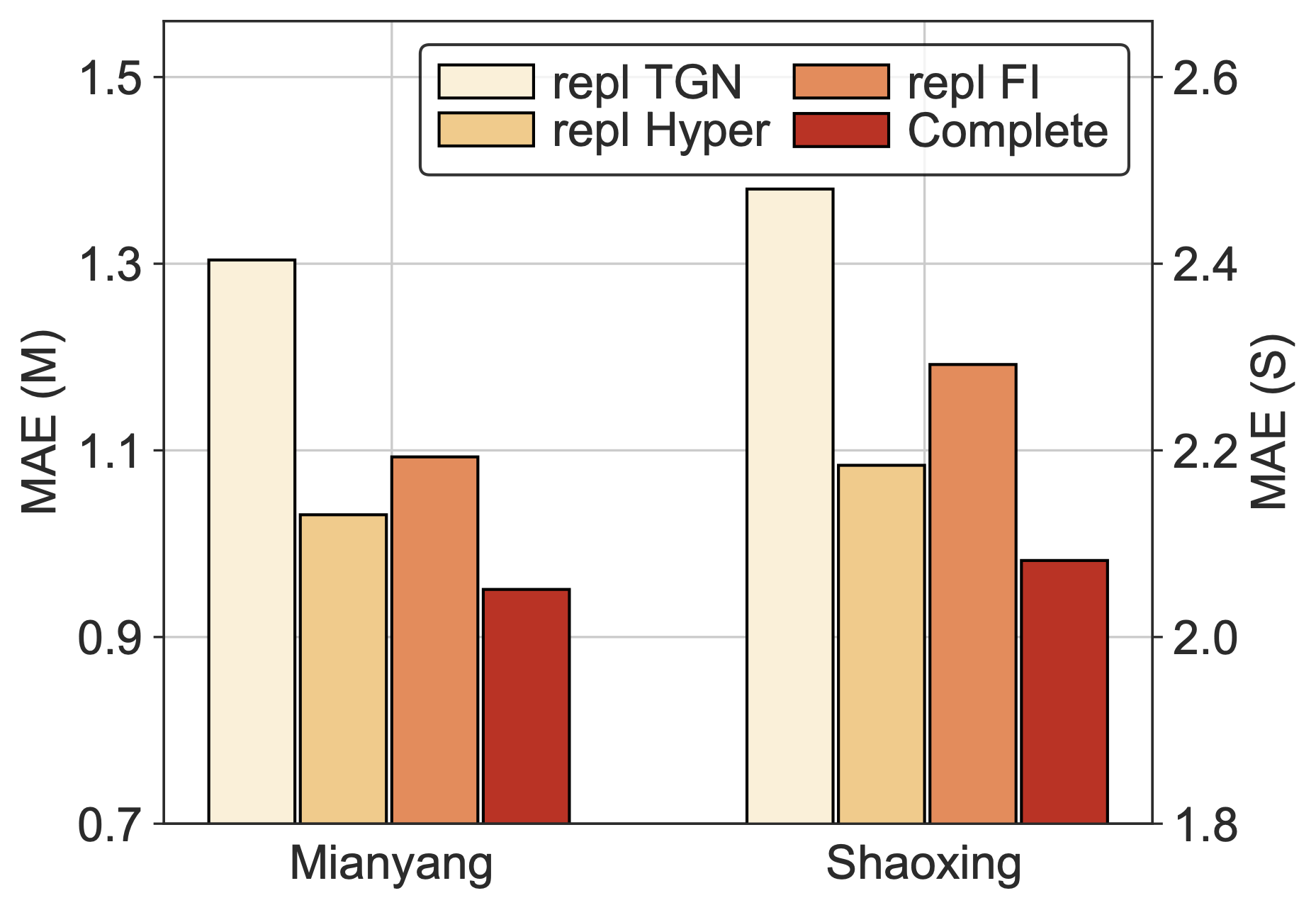}}\hspace{-1mm}
  \subfigure[{MAPE}]{
    \includegraphics[width=0.295\columnwidth]{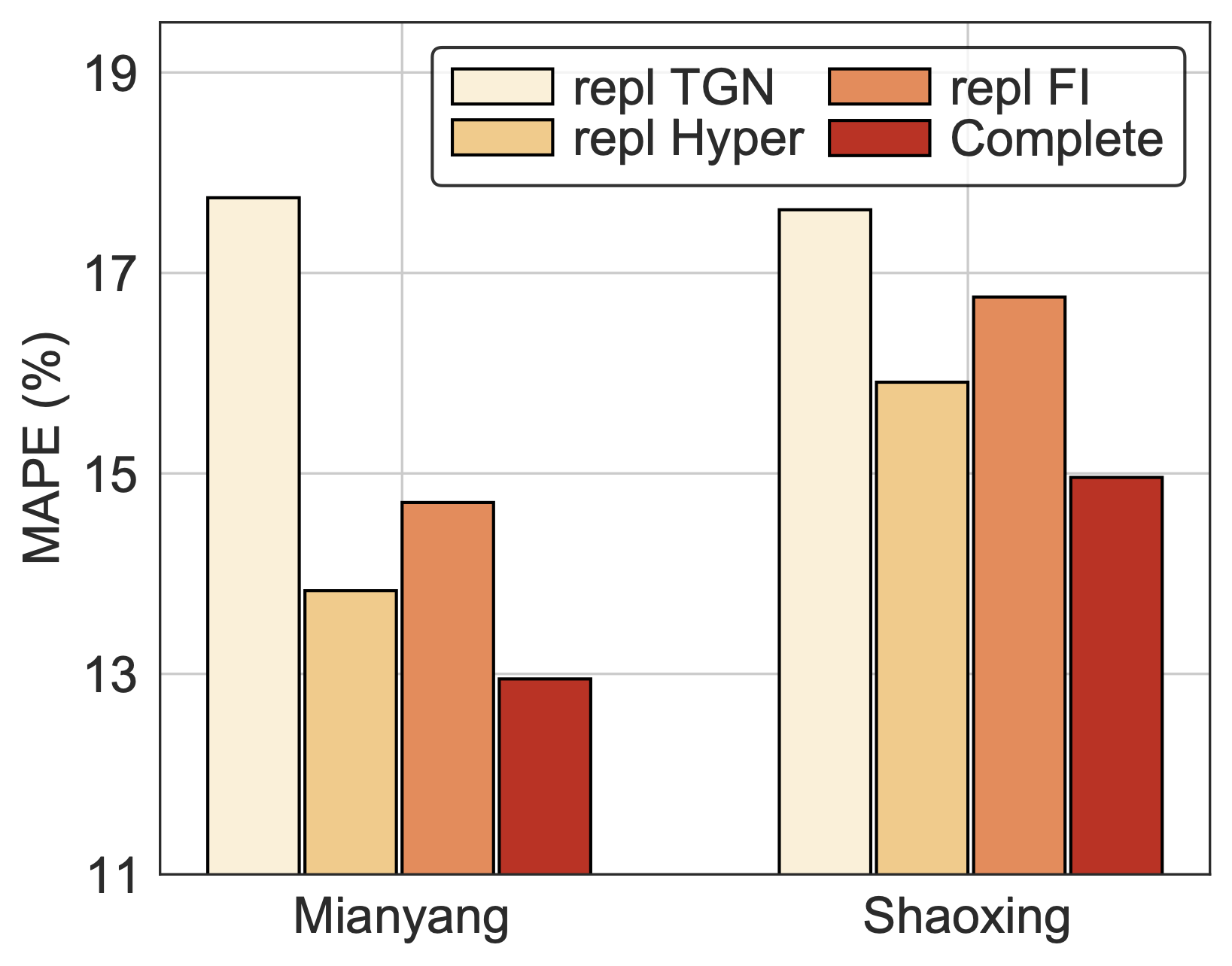}}\hspace{-0.5mm}
  \subfigure[{RMSE}]{
    \includegraphics[width=0.336\columnwidth]{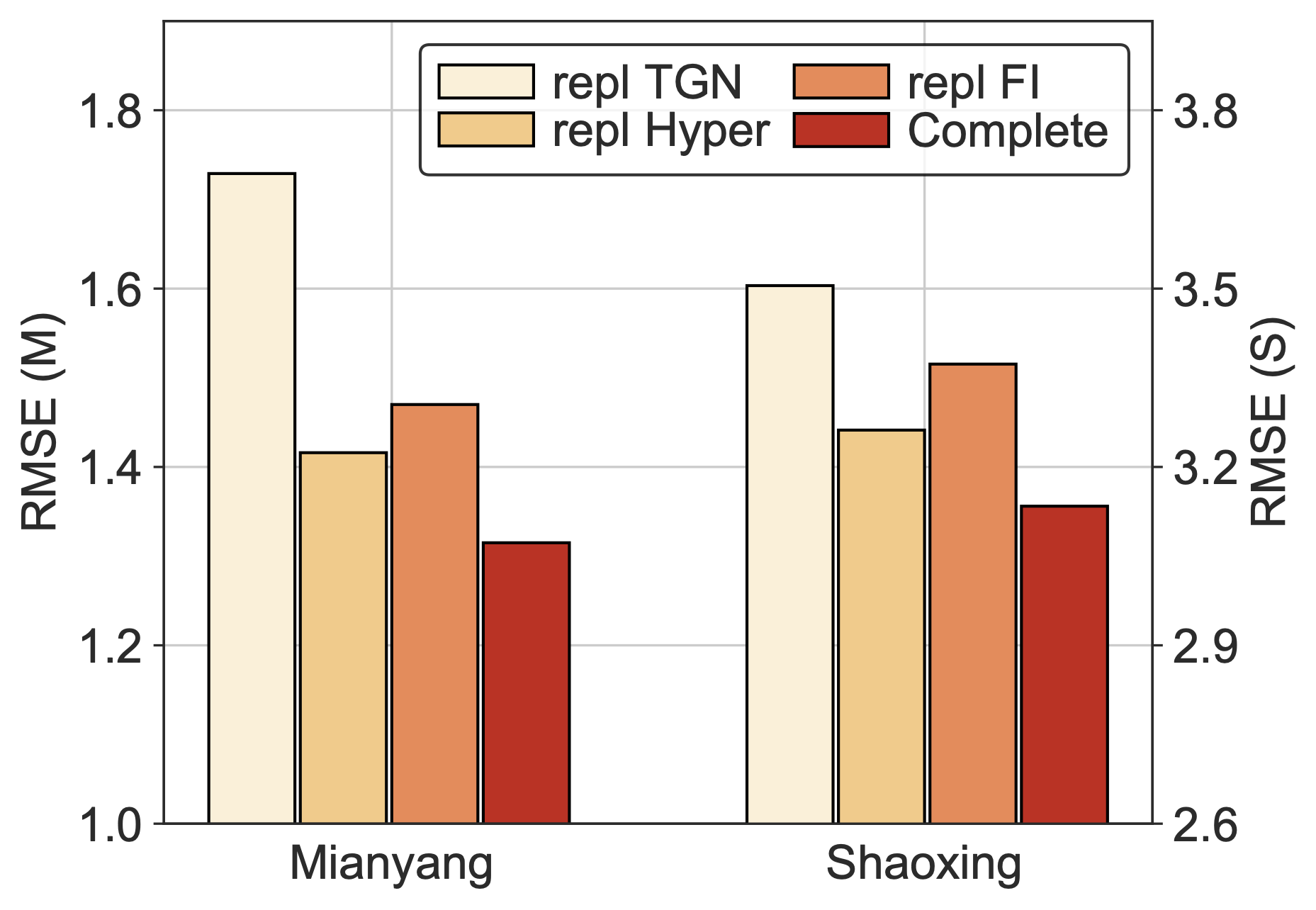}}
  \vspace{-3mm}
  \caption{Effect of transferred knowledge from a parameters perspective.} 
  \vspace{-3mm}
  \label{fig:parameter_perspective}
\end{figure}

\subsection{Effect of Different Source City}
We study the effect of different source cities on the knowledge transfer to target cities by removing the corresponding source city during meta-training. We present MAE of \meta on \mianyang and \shaoxing with 100 training instances in \figref{fig:city_effect}(a). As can be seen, for target city \mianyang, the model without source city \chengdu causes the most obvious performance degradation, while the effect is relatively slight by removing source city \guangzhou. We can explain this by \figref{fig:city_effect}(b), which shows that \mianyang is an inland city that is geographically close to \chengdu, thus they may be more related~\cite{tobler1970computer}, while \guangzhou is a coastal city and they are geographically far apart.
For target city \shaoxing, we observe the model without \guangzhou has larger performance descent than removing other source cities. This may because \shaoxing is more similar to \guangzhou as a coastal city, thus more valuable knowledge can be transferred from \guangzhou to \shaoxing.

% \vspace{-3mm}
\subsection{Effect of Different Transferred Knowledge}
We study the effects of different transferred knowledge from features perspective and parameters perspective, and present results of \meta on \mianyang and \shaoxing with 100 training instances \wrt three metrics in \figref{fig:feature_perspective} and \figref{fig:parameter_perspective}.

% \subsubsection{\textbf{Features Perspective}}
\noindent \textbf{Features Perspective.}
We first study the effect of transferable knowledge to extract different types of features via masking the corresponding features of source cities data, so that the model cannot learn how to model these features during meta-training:
% \begin{itemize}
(1)~\textbf{mask Estate} masks real estate apartment attributes;
(2)~\textbf{mask Community} masks the residential community attributes;
(3)~\textbf{mask Price} masks the historical transaction's unit price;
(4)~\textbf{Complete} doesn't mask any feature.
% \end{itemize}
As shown in \figref{fig:feature_perspective}, we can observe lack of any features knowledge results in remarkable performance decline, which verifies that all these features knowledge are positively transferable for cross-city real estate appraisal.
In addition, we find \textbf{mask Price} leads to more obvious performance reduction than \textbf{mask Estate} than \textbf{mask Community}, which indicates correlated transaction's price modeling is the most helpful knowledge to be transferred across cities by our model, then is real estate features and community features. This discovery indicates that spatiotemporal correlations between real estate prices widely exist in various cities, and how to model such price correlations is critical knowledge for real estate appraisal.

% \subsubsection{\textbf{Parameters Perspective}}
\noindent \textbf{Parameters Perspective.}
To study the effect of different transferred knowledge preserved in model's parameters, we replace a part of model's parameters that have been well trained in source cities data with the random initialization, then adapt the modified model to target city:
% \begin{itemize}
(1)~\textbf{repl TGN} replaces parameters of TGN module;
(2)~\textbf{repl Hyper} replaces parameters of hypernetworks in HMTL;
(3)~\textbf{repl FI} replaces parameters of feature integration function in \equref{equ:appraisal};
(4)~\textbf{Complete} keeps all parameters. 
% \end{itemize}
As can be seen in \figref{fig:parameter_perspective}, replacing any component's parameters leads to obvious performance degradation, which demonstrates each component of our model indeed preserves and transfers valuable knowledge from source cities to target city. Moreover, we observe replacing TGN module's parameters especially damages model's performance, which further uncovers how to effectively model the correlations between real estate transactions is the crucial knowledge for accurate real estate appraisal.

% \subsection{Speed of Adaptation}
% \subsection{Generalizability on New Community}

\subsection{Analysis on Instance Weight}
\label{app:weight}
\subsubsection{\textbf{Change of Weights}}
\begin{figure}[tb]
  % \vspace{-0.5mm}
  \centering
  % \hspace{-2mm}
  \vspace{-2mm}
  \subfigure[{Transfer to Mianyang}]{
    \includegraphics[width=0.4\columnwidth]{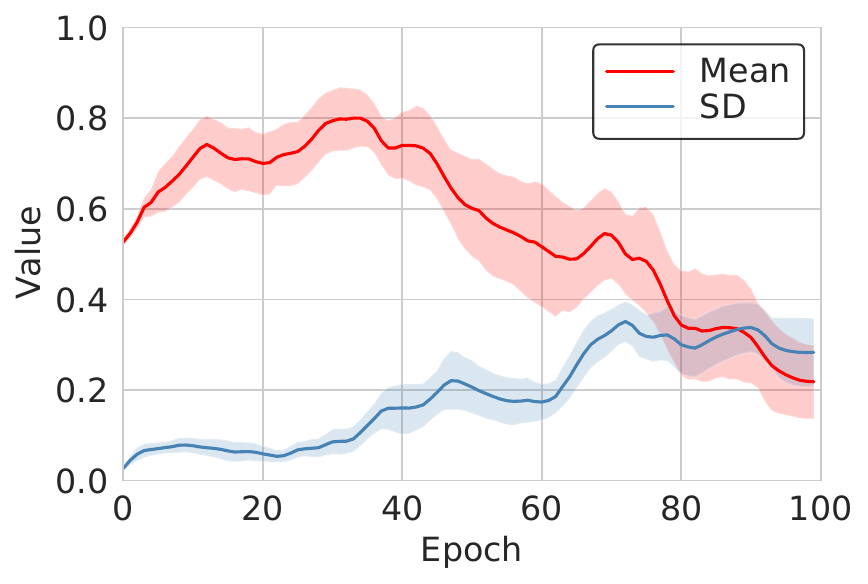}}
  \subfigure[{Transfer to Shaoxing}]{
    \includegraphics[width=0.4\columnwidth]{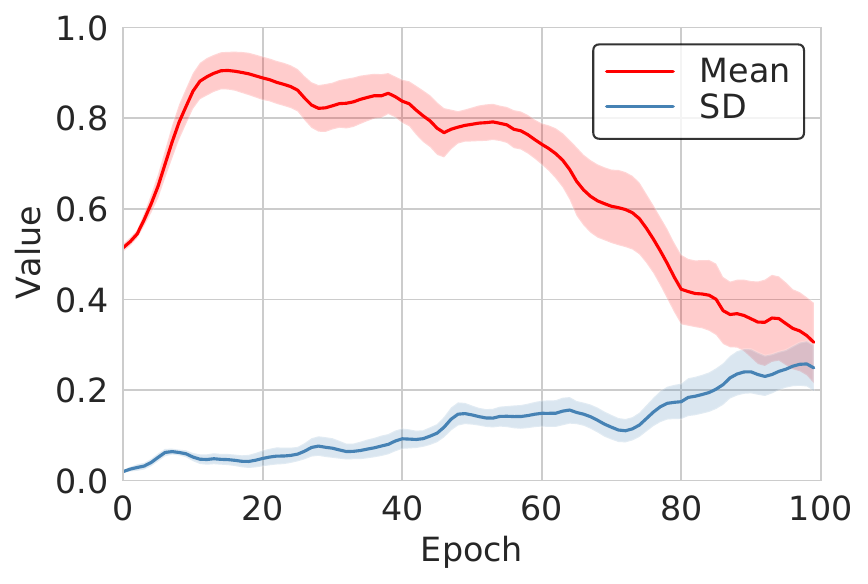}}
  \vspace{-2mm}
  \caption{Change of Weights.} 
  \vspace{-3mm}
  \label{fig:weight_change}
\end{figure}

\figref{fig:weight_change} shows the change of instance weights during meta-training for target city \mianyang and \shaoxing, where \textbf{Mean} denotes the mean of all weights for source instances, and \textbf{SD} refers to standard deviation of all weights. The meta-training is repeated 5 times with different random seeds. As can be observed, the overall trend of instance weights \textbf{Mean} value first increases, then gradually decreases. This is because at first most instances from source cities are valuable for target city, thus the instance weights \textbf{Mean} value becomes larger to enhance knowledge transfer at the beginning of meta-training. However, with meta-training proceeding, the effect of many source instances may be exhausted, or even be negative. Hence, the weights for these useless instances become small. Moreover, we observe the instance weights \textbf{SD} value always keeps an increasing trend. This is because the useful and useless source instances become divergent as meta-training goes, weight-generating network generates larger weights for these useful instances and smaller weights for useless instances. It indicates our model to mitigate negative knowledge transfer across cities.

\subsubsection{\textbf{Case Study}}
\begin{figure}[tb]
% \vspace{1mm}
  \centering
  % \hspace{-1mm}
  \includegraphics[width=0.8\columnwidth]{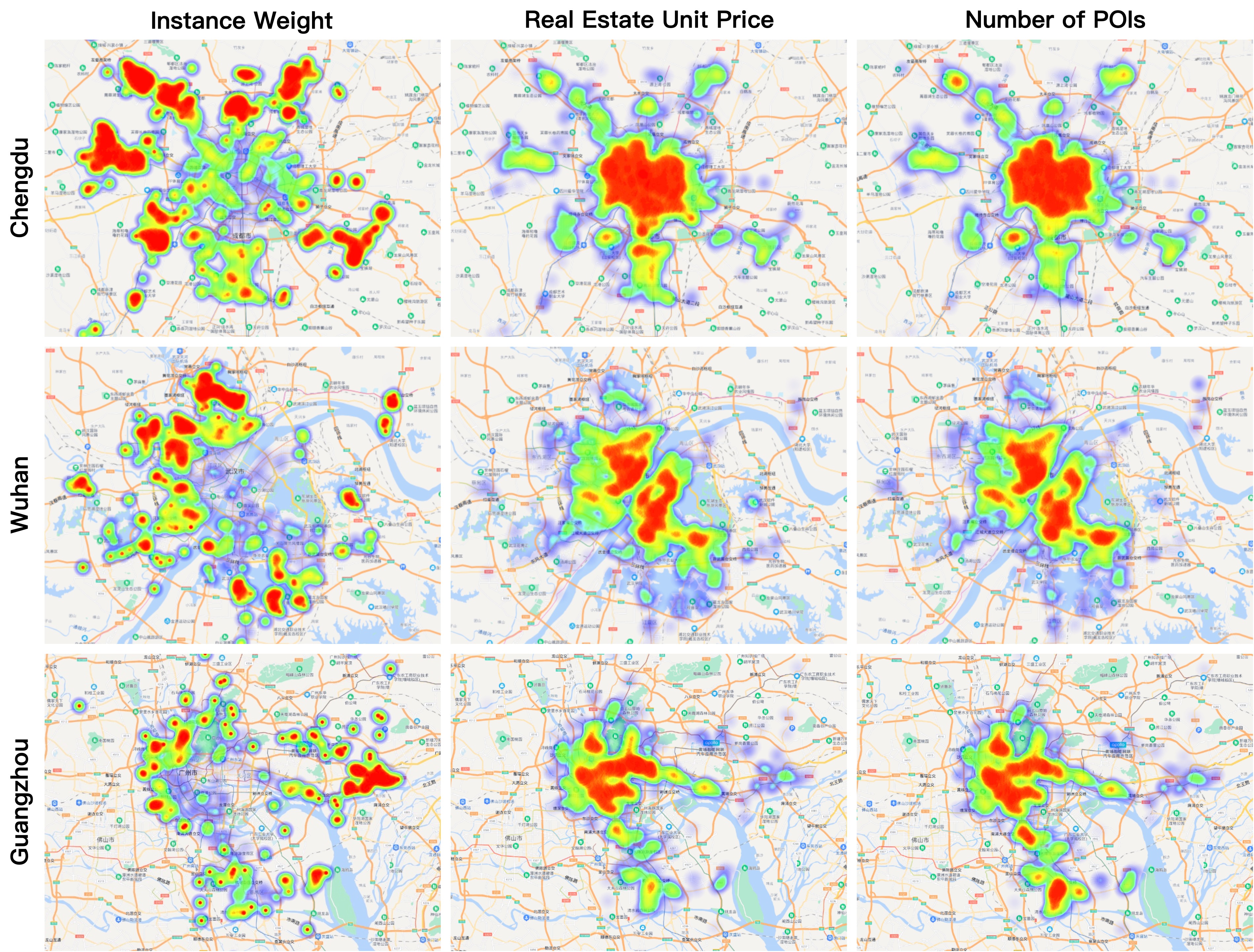}
  % \vspace{-3mm}
  \caption{Visualization for the spatial distributions of instance weight, real estate unit price, and the number of POIs on source cities. Brighter colors represent larger values.} 
  % \vspace{-3mm}
  \label{fig:heatmap}
\end{figure}

\figref{fig:heatmap} depicts the spatial distribution of source instances weights when transfer knowledge to \mianyang, and the spatial distributions of real estate prices and the number of POIs around community. 
We can discover that the instances weights are negatively correlated with real estate prices and the number of POIs. One possible explanation is lower real estate prices and less POIs usually signify unprosperous areas, and the real estates in unprosperous areas of metropolises have more similar pattern to real estates in small cities like \mianyang. Thus, our model weights more on these source instances of unprosperous areas to learn more valuable knowledge for real estate appraisal in target city.

% \vspace{-3mm}
\subsection{Efficiency Test}
\label{sec:latency}
\begin{figure}[tb]
  % \vspace{-2mm}
  \centering
  \includegraphics[width=0.7\columnwidth]{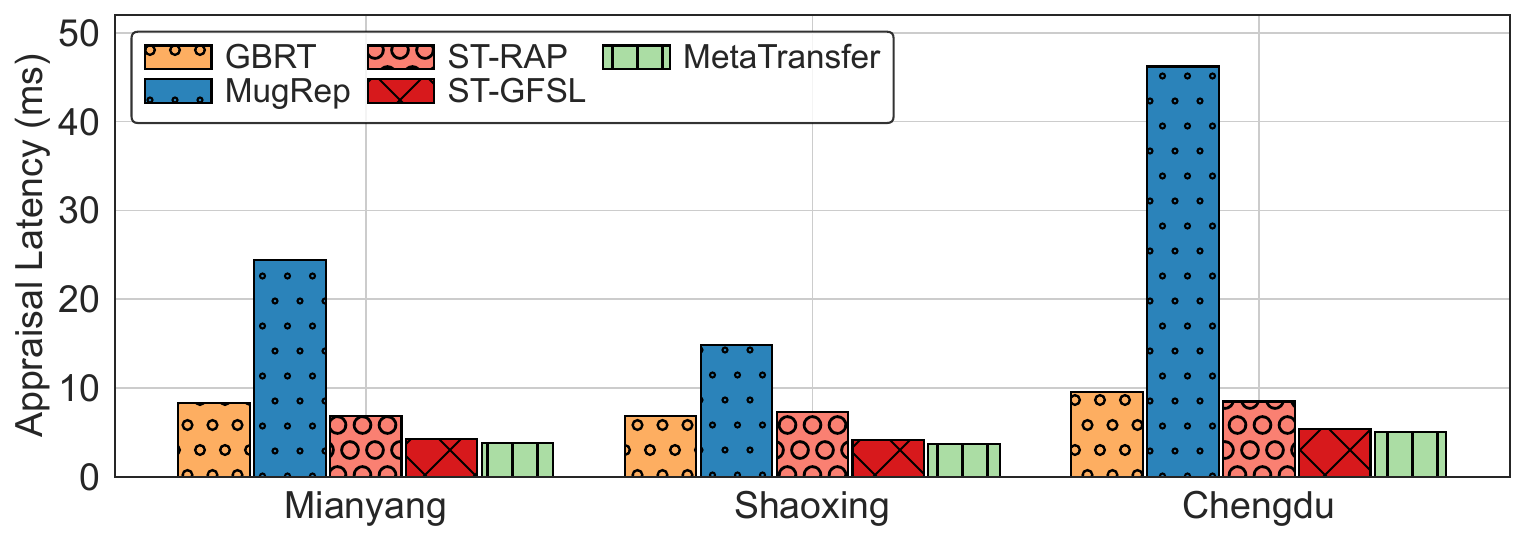}
  % \vspace{-6mm}
  \caption{Average real estate appraisal latency.} 
  % \vspace{-3mm}
  \label{fig:latency}
\end{figure}

\rev{\figref{fig:latency} displays the average latency of various models when responding to a real estate appraisal request. The results indicate that \meta~(\ie MTTGN) has the lowest response latency among these models, on average requiring 3.8ms, 3.7ms, 5.1ms for \mianyang, \shaoxing, and \chengdu, respectively. It suggests that \meta has a highly efficient inference process for practical use, even for metropolis with vast data instances. 
\meta has superior efficiency than GBRT (8.3ms, 6.9ms, 9.6ms) for it does not require a time-consuming model ensemble.
Particularly, compared to MugRep (24.4ms, 14.9ms, 46.2), \meta is notably more efficient because it has efficiently optimized the correlation modeling process between real estate transactions, while MugRep repeatedly involves all neighboring historical transactions into computation to respond to each real estate appraisal request, leading to substantially increasing computational overhead as the number of correlated transactions grows.
Moreover, \meta is more efficient than ST-RAP, which requires multiple graph convolution operations for its diverse neighboring entities, such as communities, amenities, and transportation stations.
All these results demonstrate the efficiency, scalability, and practicality of \meta.}

% \vspace{-1mm}
% \input{Related}
\section{Related work}\label{sec:related}
% In this section, we present a review of prior related literature on real estate appraisal and cross-city knowledge transfer.

% \subsection{Real Estate Appraisal}
\noindent \textbf{Real Estate Appraisal.}
Traditional valuation methods for real estate primarily include income approach, cost approach, sales comparison approach, and hedonic price model.
Income approach~\cite{baum2017income} appraises real estate market value based on the income that real estate can generate. Cost approach~\cite{guo2014integrated} appraises real estate by considering the land and construction values as well as the improvements' depreciated cost. Sales comparison approach~\cite{Mccluskey1997AnEO} estimates real estate market value relying on comparison with recent sales of similar real estates. Hedonic price model~\cite{cheshire1995price} assumes that the real estate market value can be regarded as the aggregation of its individual characteristics.
However, these methods heavily rely on expert knowledge~\cite{pagourtzi2003real}, which prevents their application by non-experts.
% However, these approaches heavily rely on the availability, accuracy, timeliness of sale transaction data and domain knowledge~\cite{pagourtzi2003real}, and fail to consider attributes interactions which leads them to be incompetent in complex non-linear data~\cite{limsombunchai2004house}.
Besides, automated valuation methods appraise real estate market value based on automatic machine learning techniques~\cite{valier2020performs}, such as linear regression~\cite{csipocs2008linear}, support vector regression~\cite{lin2011predicting}, boosted regression trees~\cite{park2015using}, and deep neural networks~\cite{you2017image,law2019take,zhang2021mugrep,du2023dora}. These methods have garnered significant interest among researchers because they enable automated real estate appraisal, making them easily accessible even to non-domain experts, and thus are with excellent practicability~\cite{niu2019intelligent}.
Furthermore, some works~\cite{fu2014exploiting, fu2014sparse, fu2015real} study to rank real estates by considering the perspectives of mixed land use, online user reviews and offline moving behaviors, as well as real estate individual, peer and zone dependency.
In addition, there are a few studies attempting to incorporate the relations between real estates.
% that try to capture the peer-dependency among nearby estates.
% \citet{fu2014exploiting} use the generative likelihood of each edge to model peer-dependency, which does not adequately integrate the attributes knowledge of nearby estates.
For example, You et al.~\cite{you2017image}~use a random walk strategy to generate real estate sequences based on location, followed by a recurrent neural network for price prediction, but this sampling method may result in the loss of valuable information.
The recent works~\cite{zhang2021mugrep,lee2023st}~introduce graph neural networks to model the dependencies of real estate transactions and residential communities. 
However, they fail to consider time-aware correlations among transactions and the diversity of communities. 
% Moreover, the model in \cite{zhang2021mugrep} is much more computationally expensive~(see \secref{sec:latency}) as it needs to repeatedly involve all neighboring historical transactions for each real estate appraisal response.
Most importantly, all the above studies investigate real estate appraisal for a single city dataset. The cross-city real estate appraisal problem remains under-explored.

% \subsection{Cross-City Knowledge Transfer} 
\rev{\noindent \textbf{Cross-City Knowledge Transfer.} 
Previous studies for cross-city knowledge transfer can be categorized into single source city transfer learning~\cite{wei2016transfer,wang2019cross,ijcai2022p282,guo2018citytransfer,jin2022selective,ouyang2023citytrans,jin2023transferable} and multiple source cities transfer learning~\cite{yao2019learning,luCrossCityTransfer22,yuan2024spatio} in terms of the source cities number. 
As an early single source city transfer learning method, Wei et al.~\cite{wei2016transfer} aims to improve air quality prediction by transferring the learned semantically related dictionaries and labelled instances from a source city to a target city. 
Wang et al.~\cite{wang2019cross} studies to transfer the region-level features between cities by learning an inter-city region matching function to match each region of a target city to a similar region in a source city. 
% Fang et al.~\cite{ijcai2022p282} adopts a spatial adversarial adaptation module with a temporal attentive adaptation module to simultaneously enable spatiotemporal knowledge transfer between cities in urban flow prediction task. 
% Guo et al.~\cite{guo2018citytransfer} solves the cold-start problem of chain store site recommendation in a new city based on the collaborating filtering and chain store knowledge transfer from a data-rich city. 
Jin et al.~\cite{jin2022selective} adopts a meta-learning paradigm to train a region weighting network, then the prediction model can be pre-trained on a single source city with learned region weights to initialize fine-tuning on the target city. The training strategy of our weight-generating network is connected to \cite{jin2022selective}, but we present a tri-level optimization framework to realize more fine-grained instance-level re-weighting on multiple source cities data for the meta-training process of meta-learning~\cite{finn2017model} rather than traditional pre-training adopted in \cite{jin2022selective}. 
However, transferring knowledge from only a single source city tends to cause unstable and useless transfer risks~\cite{yao2019learning}. 
Hence, a few recent works~\cite{yao2019learning,luCrossCityTransfer22,yuan2024spatio} introduce meta-learning or diffusion models aiming to transfer knowledge from a set of tasks extracted from multiple source cities to a new task of the target city.
% For example, studies~\cite{yao2019learning, luCrossCityTransfer22} introduce meta-learning to achieve cross-city knowledge transfer for traffic prediction problem by regarding each city as a task. 
% Different from them, in our problem, each city is formulated as a task set~(\ie~a set of tasks), then the knowledge is transferred between different task sets.
Nevertheless, existing works primarily focus on studying cross-city knowledge transfer in the context of spatiotemporal prediction problems~(\eg~traffic prediction~\cite{liu2022practical,yuan2025universal,lei2025st}) with regular-interval time series data, preventing them from being directly applied to both spatially and temporally irregular data, such as real estate transaction events ~\cite{zhangirregularkdd24,zhangirregularicml24} and trajectories~\cite{fang2023ghost,gao2023efficient,wu2024trajrecovery}.
Additionally, they fail to consider the dynamic effect of each source instance on knowledge transfer to the target city, which may encounter the risk of negative transfer.
}

% \vspace{-1mm}
% \input{Conclusion}
\section{Conclusion}\label{sec:conclusion}
In this paper, we investigated the cross-city real estate appraisal problem which aims to transfer valuable knowledge from multiple data-rich source cities to the data-scarce target city to improve valuation performance. Specifically, we proposed Meta-Transfer Learning Powered Temporal Graph Networks, \meta, where the task of city-wide real estate appraisal was reformulated as a multi-task dynamic graph link label prediction problem. 
Along this line, we first designed an Event-Triggered Temporal Graph Network to model the
irregular spatiotemporal correlations among evolving real estate transactions.
Then, a Hypernetwork-Based Multi-Task Learning module was proposed to simultaneously facilitate intra-city knowledge sharing and community-wise knowledge learning. 
Furthermore, to achieve effective cross-city knowledge transfer, we proposed a Tri-Level Optimization Based Meta-Learning framework to adaptively re-weight training transaction instances from multiple source cities so that transfer helpful knowledge to the target city.
\rev{Finally, extensive experiments on six real-world datasets demonstrated the effectiveness of \meta.}

% \section*{Acknowledgments}
\begin{acks}
This work was supported in part by the National Key R\&D Program of China (Grant No.2023YFF0725-001),in part by the National Natural Science Foundation of China (Grant No.92370204, No.62572417),in part by the guangdong Basic and Applied Basic Research Foundation (Grant No.2023B1515120057),in part by the Key-Area Special Project of Guangdong Provincial Ordinary Universities (2024ZDZX1007).
\end{acks}

\bibliographystyle{ACM-Reference-Format}
\balance
\bibliography{ref}

% \appendix
% \clearpage
% \newpage

% \section{Additional Experiments}

% \begin{figure}[tb]
%   \centering
%   \vspace{2mm}
%   \includegraphics[width=0.95\columnwidth]{figs/subproblem.png}
%   \caption{City-Wide Multi-Task Dynamic Graph Link Label Prediction Problem.} 
%   \label{fig:subproblem}
% \end{figure}

% \begin{figure}[tb]
%   \centering
%   \includegraphics[width=1.0\columnwidth]{figs/TGN.png}
%   \caption{Event-Triggered Temporal Graph Network. It encompasses a time-aware embedding evolution module to incorporate evolving transaction events for the embedding update and a dimensional attentive graph convolution to integrate neighboring nodes for the embedding aggregation.}
%   \label{fig:tgn}
% \end{figure}

\end{document}